\documentclass[10pt,twocolumn,letterpaper]{article}

\usepackage[pagenumbers]{cvpr} 

\definecolor{cvprblue}{rgb}{0.21,0.49,0.74}
\usepackage[pagebackref,breaklinks,colorlinks,allcolors=cvprblue]{hyperref}

\usepackage{amsmath,amsfonts,bm}

\def\eqref#1{equation~\ref{#1}}

\def\1{\bm{1}}

\DeclareMathAlphabet{\mathsfit}{\encodingdefault}{\sfdefault}{m}{sl}
\SetMathAlphabet{\mathsfit}{bold}{\encodingdefault}{\sfdefault}{bx}{n}

\definecolor{iccvblue}{rgb}{0.21,0.49,0.74}
\definecolor{nvgreen}{RGB}{118, 185, 0}
\usepackage{url}
\usepackage[capitalize]{cleveref}
\usepackage{booktabs}       
\usepackage{amsfonts}       
\usepackage{nicefrac}       
\usepackage{microtype}      
\usepackage{graphicx}
\usepackage{enumitem}
\usepackage{tcolorbox}
\usepackage{listings}
\usepackage[table]{xcolor}         
\usepackage{multirow}
\usepackage{xspace}
\usepackage{algorithm}
\usepackage{algorithmic}

\title{VideoITG: Multimodal Video Understanding with \\ Instructed Temporal Grounding}

\author{
Shihao Wang$^{1}$\thanks{Work done during an internship at NVIDIA.}~\,,
Guo Chen$^{2*}$,
De-An Huang$^3$,
Zhiqi Li$^{2*}$,
Minghan Li$^4$,
Guilin Liu$^3$,\\
Jose M. Alvarez$^3$,
\textbf{Lei Zhang}$^1$\thanks{Equal corresponding and advising authors.}~\,,
\textbf{Zhiding Yu}$^{3\dagger}$\\[0.25cm]
$^1$The Hong Kong Polytechnic Univ.~~~$^2$Nanjing Univ.~~~$^3$NVIDIA~~~$^4$Harvard Univ.\\
\href{https://nvlabs.github.io/VideoITG/}{https://nvlabs.github.io/VideoITG/}
}

\begin{document}
\maketitle
\begin{abstract}

While Video Large Language Models (Video-LLMs) have shown significant potential in multimodal understanding and reasoning tasks, how to efficiently select the most informative frames from videos remains a critical challenge. Existing methods attempt to optimize frame sampling by reducing inter-frame redundancy or employing unsupervised event localization. However, these approaches often fall short in handling complex instruction-following tasks and scenarios that demand precise temporal modeling, resulting in limited performance in both semantic alignment and temporal reasoning. To address the above challenges, we introduce \textit{Instructed Temporal Grounding for Videos} (\textbf{VideoITG}), a framework aiming to adaptively customize frame sampling strategies based on user instructions. Specifically, we design the VidThinker pipeline, which automates annotation by generating instruction-conditioned captions, retrieving relevant video segments, and selecting key frames to enable efficient supervision. Using VidThinker, we build the VideoITG-40K dataset with 40K videos and 500K temporal grounding annotations. Our plug-and-play VideoITG model leverages Video-LLMs’ visual-language alignment and reasoning for discriminative frame selection. VideoITG consistently boosts the performance on multiple multimodal video understanding benchmarks, demonstrating its effectiveness and potential.

\end{abstract}

\section{Introduction}
\begin{figure}[!h]
\centering
\includegraphics[width=0.95\linewidth]{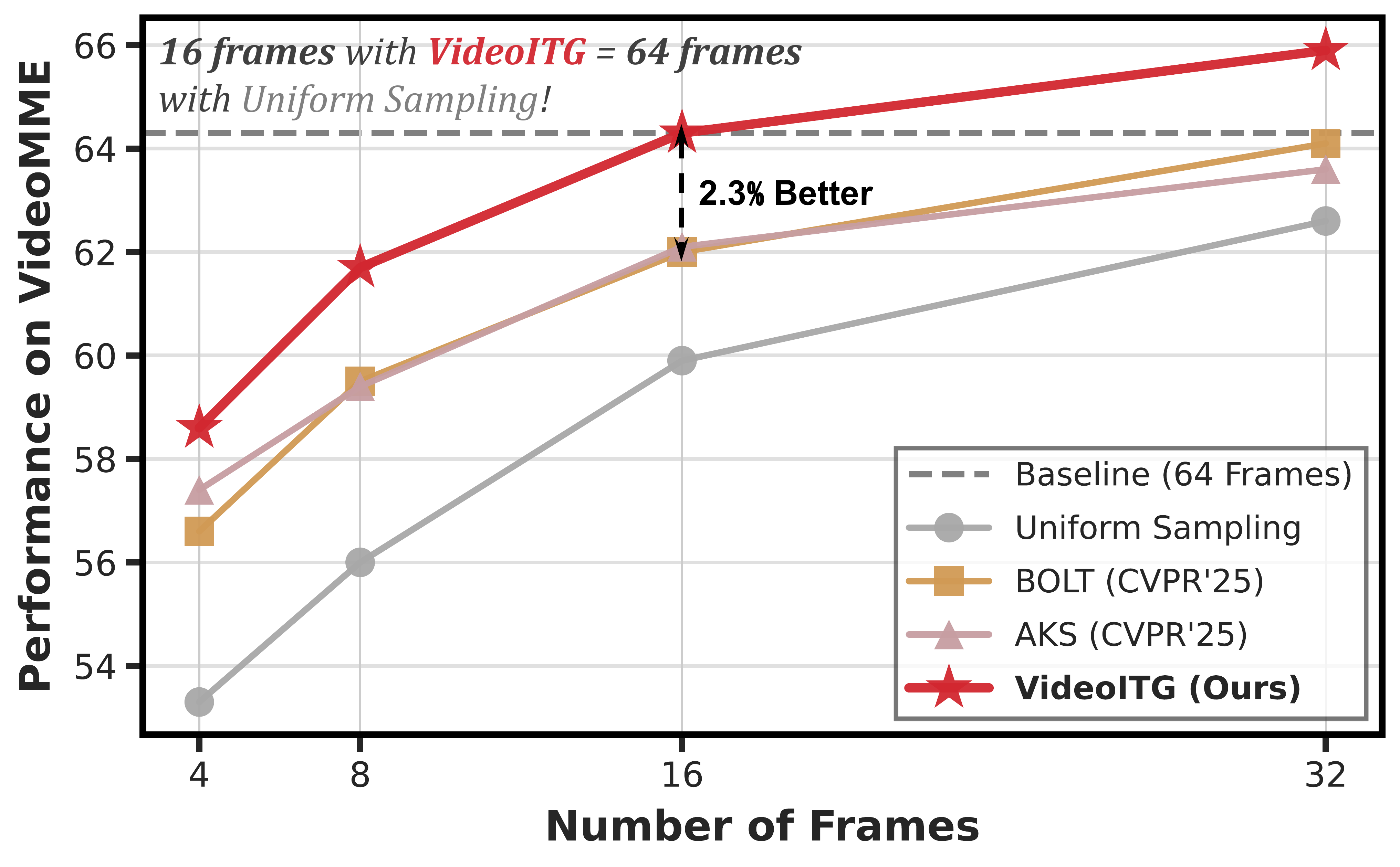}
\vspace{-3mm}
\caption{\noindent \textbf{Comparison of different frame selection methods on VideoMME with LLaVA-Video-7B.} VideoITG consistently enhances baseline methods and achieves state-of-the-art performance.}
\label{fig:frames}
\vspace{-2mm}
\end{figure}
The rapid progress of Video Large Language Models (Video-LLMs) has opened new frontiers in video understanding, enabling complex tasks such as captioning~\cite{chensharegpt4video, chai2025auroracap, zhou2024streaming, islam2024video, chen2024large, wang2024tarsier}, visual question answering~\cite{bai2025qwen3, fu2024videomme, zhou2024mlvu, mangalam2024egoschema, li2024mvbench, chen2024cg, xiao2021next, patraucean2023perception}, and even embodied-agent interaction~\cite{brohan2023rt, kim2024openvla, fu2024vita, liu2024streamchat, chen2024videollm, chen2025grounded}. 
However, these models still struggle with long videos, where high memory cost and computation overhead limit their ability to process extended temporal contexts. 
A common workaround is uniform frame sampling—simple yet suboptimal—often missing key frames critical for semantic and temporal reasoning, thereby constraining overall performance.

To alleviate this challenge, prior studies have explored multiple directions. 
One class of approaches focuses on reducing spatiotemporal redundancy through pooling~\cite{xu2024pllava, shen2024longvu}, similarity pruning~\cite{zhang2024flash}, or clustering-based compression~\cite{li2024llavaonevision, zhang2024internlm}, retaining only essential frames. 
Another line extends model sequence length to capture long-term dependencies~\cite{wang2024qwen2, team2023gemini}, yet such strategies incur high computation and risk information dilution. 
Other methods incorporate question-centric cues for frame selection~\cite{li2024llama, yu2023self}, demonstrating superiority over uniform sampling (\ref{fig:frames}). 
For example, SeViLA~\cite{yu2023self} applies BLIP-2~\cite{li2023blip} to process each frame independently before selecting keyframes, which are then fed into video reasoning pipelines. 
Nevertheless, the absence of temporal modeling across frames hinders effective reasoning over multi-event or time-sensitive queries.

\begin{figure*}[t!]
\centering
\vspace{-1mm}
\includegraphics[width=0.92\linewidth]{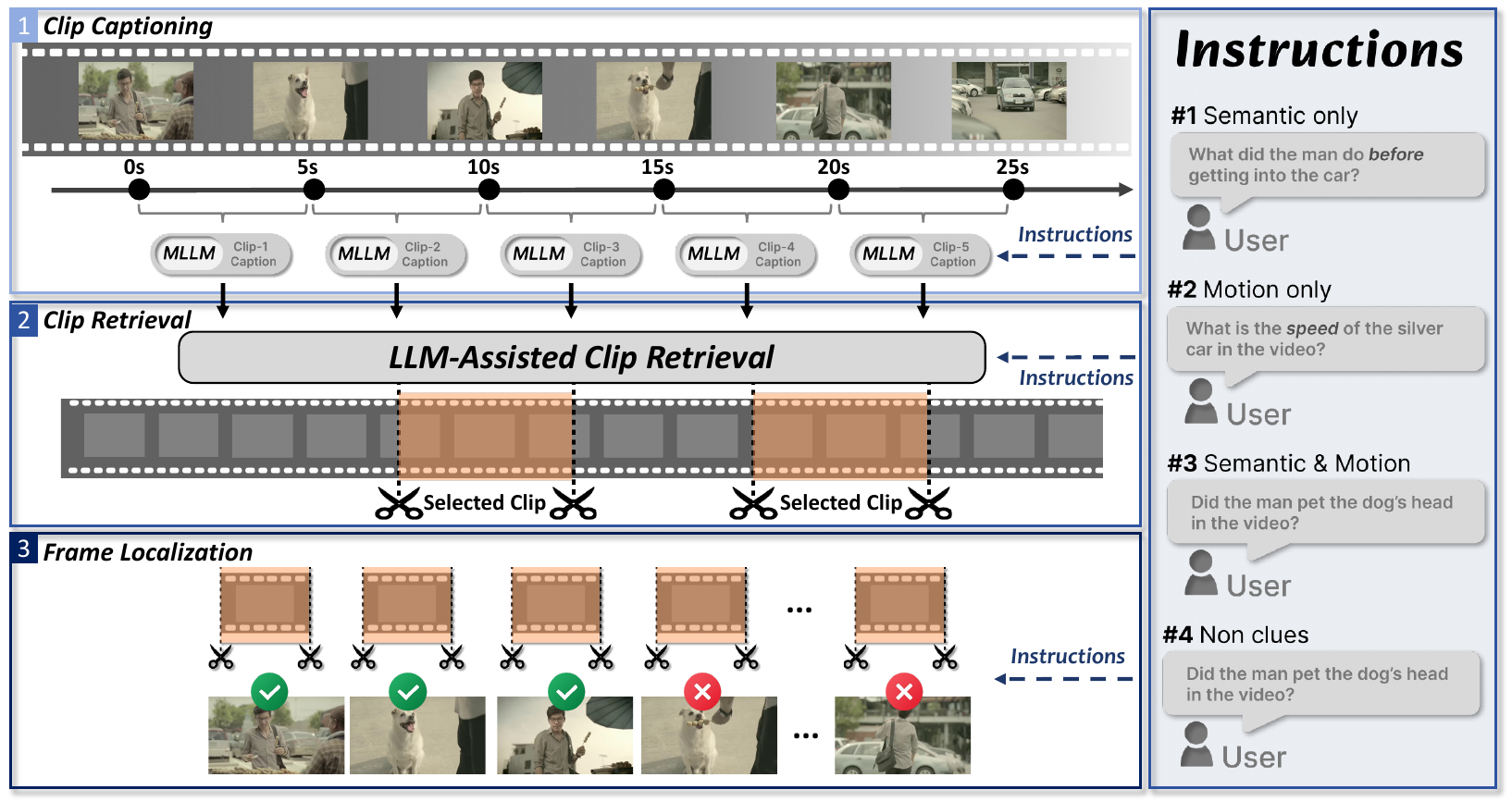}
\vspace{-2mm}
\caption{
\textbf{Overview of the \textit{VidThinker} annotation pipeline for VideoITG.}
It consists of three human-inspired stages: (1) clip-level captioning under instructions; (2) instruction-guided relevant clip retrieval; and (3) fine-grained frame-level localization.
}
\label{intro}
\vspace{-4mm}
\end{figure*}

Despite progress in compressing or extending temporal contexts, a substantial performance gap remains between short and long videos—primarily due to the lack of large-scale, instruction-guided temporal grounding data. 
When humans analyze long videos, they rarely process all frames at once; instead, they skim for global context, identify question-relevant cues, and zoom in on discriminative moments. 
Inspired by this human strategy, we propose \textbf{Instructed Temporal Grounding for Videos (VideoITG)}, which integrates user instructions directly into the frame selection process. 
While traditional temporal grounding~\cite{wang2024grounded, qian2024momentor, lei2021detecting} focuses on localizing events using single descriptive queries, \textbf{VideoITG} introduces instruction-driven temporal reasoning, adaptively customizing the sampling strategy for each task. 
Unlike prior frame selection frameworks~\cite{yu2023self, yu2025frame, han2025videoespresso, meng2022adavit, wang2022efficient}, 
our method handles multi-temporal and multi-cue scenarios by (i) localizing temporal cues across clips for relational reasoning, (ii) employing hybrid sampling for dynamic event variations, and (iii) maintaining holistic coverage for content verification and captioning.

To support VideoITG, we construct a large-scale dataset via an automated annotation pipeline named \textit{VidThinker}. 
As shown in Fig.~\ref{intro}, VidThinker automates data generation through instruction-conditioned clip captioning, instruction-guided retrieval, and fine-grained frame localization. 
Driven by GPT-4o~\cite{openai2024gpt4o} reasoning, VidThinker emulates a "Needle-in-a-Haystack" process to retrieve relevant moments and provides balanced supervision across four instruction types: 
\textbf{(1) semantic-only}, focusing on appearance; 
\textbf{(2) motion-only}, emphasizing dynamic cues; 
\textbf{(3) semantic \& motion}, for joint reasoning; and 
\textbf{(4) non-clues}, open-ended video-level prompts that require maximizing visual diversity across the entire video.

The resulting \textbf{VideoITG-40K} dataset contains 40K videos (30s–3min) and 500K instruction-grounded annotations—surpassing existing temporal grounding datasets by more than \textbf{4$\times$} in both scale and instruction quality. 
Building on this foundation, we design a family of VideoITG models—featuring text generation, anchor-based causal attention, and full-attention pooling—to effectively align temporal cues with user instructions. 

In summary, our key contributions are as follows:
\begin{itemize}[leftmargin=6mm, itemsep=3pt, parsep=0pt, topsep=2pt]
    \item \textbf{VideoITG-40K dataset.} 
    Built via the automated \textit{VidThinker} pipeline, we curate \textbf{VideoITG-40K}, it contains 40K videos and 500K fine-grained, instruction-aligned annotations, substantially expanding the scale and diversity of existing temporal grounding resources.
    \item \textbf{VideoITG models.} 
    We propose three complementary model variants that explore distinct attention and decoding mechanisms, offering a unified, plug-and-play framework adaptable to diverse Video-LLMs.
    \item \textbf{Consistent improvement.} 
    Across benchmarks and models, VideoITG boosts accuracy with fewer frames: on VideoMME (Fig.~\ref{fig:frames}), 16 frames with VideoITG match 64-frame uniform sampling and are comparable to SOTA methods using 32 frames.
\end{itemize}

\section{Related work}

\begin{figure*}[ht]
\vspace{-2mm}
\centering
\includegraphics[width=0.96\linewidth]{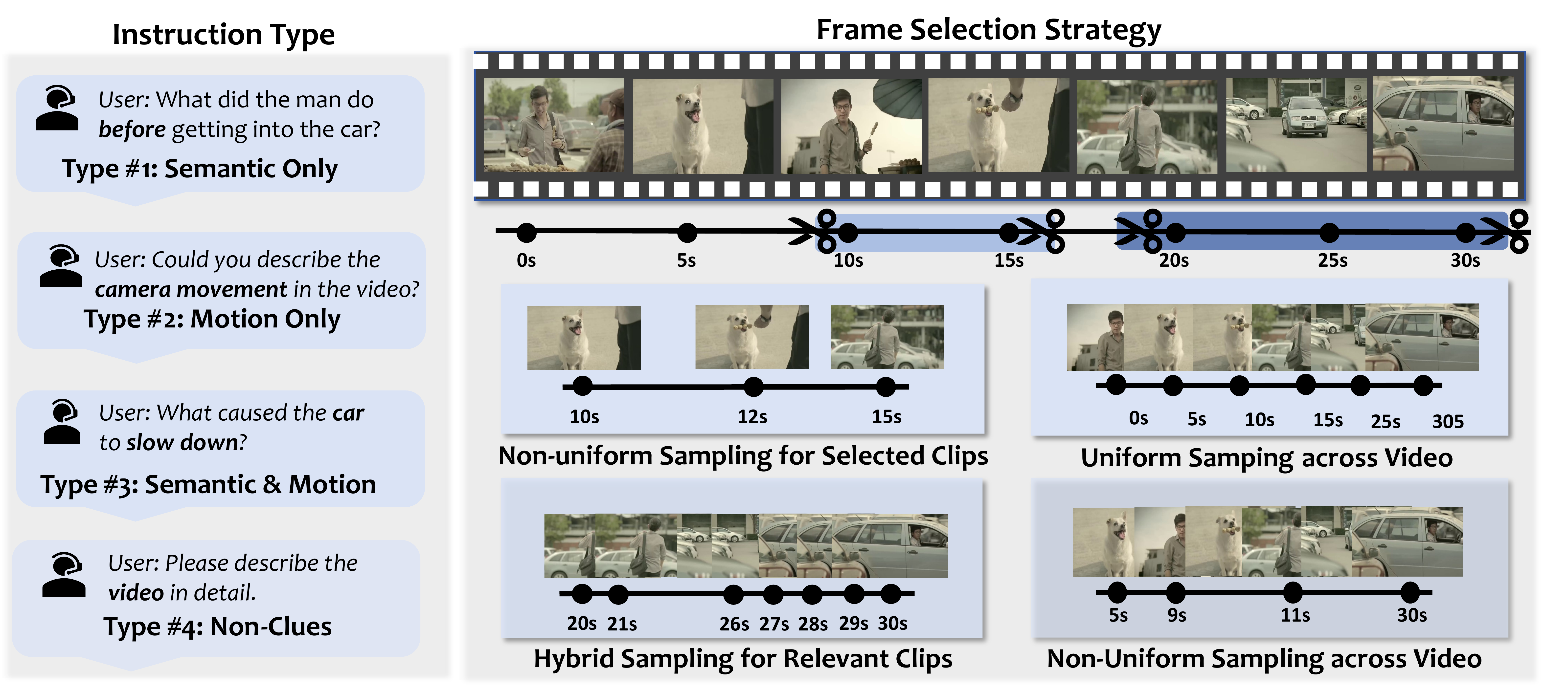}
\vspace{-2mm}
\caption{\textbf{Illustration of four instruction types and their corresponding frame selection strategies in VidThinker.} For semantic-focused instructions, the system selects diverse frames capturing key visual clues. For motion-focused instructions, frames are uniformly sampled to capture dynamic changes. When both semantic and motion cues are required, a hybrid sampling strategy is applied. For vague or open-ended instructions, the system samples a minimal yet diverse set of frames across the video for holistic coverage.}
\label{dataset}
\vspace{-2mm}
\end{figure*}

\noindent \textbf{Video large language models.} Recent advances in Video-LLMs address the temporal and spatial complexity of long videos through several strategies. Visual feature compression~\cite{liu2025oryx, zohar2024apollo, ye2024mplug, wang2024videollamb, liu2025video} is achieved by modules like Q-Former~\cite{song2024moviechat} and Perceiver Resampler~\cite{zohar2024apollo}, which merge frame features into fixed queries. Spatial pooling~\cite{maaz2024video, xu2024slowfast, xu2024pllava} helps preserve long-range temporal information efficiently. Some models extend sequence length for longer inputs~\cite{zhang2024long, wang2024qwen2, team2023gemini, chen2025eagle}, but this often increases computational cost~\cite{wei2024visual, shu2025video}. To reduce redundancy, similarity-based frame filtering is used~\cite{jin2024video, shen2024longvu}, though fixed thresholds may miss real-world diversity.

\noindent \textbf{Keyframe selection for Video-LLMs.} The goal of keyframe selection is to choose a compact subset of frames that preserves task-critical semantics and temporal evidence while maintaining sufficient video-level coverage for long-form understanding. Recent methods increasingly incorporate user instructions to rank or retrieve frames most relevant to the query and then feed only these keyframes to the downstream Video-LLM~\cite{liu2025boltboostlargevisionlanguage,tang2025adaptive,luo2025quota,yu2025framevoyagerlearningqueryframes,zhang2025q}.
AKS~\cite{tang2025adaptive} performs adaptive sampling guided by temporal characteristics to reduce redundancy without sacrificing content coverage.
Q-Frame~\cite{zhang2025q} introduces a query-adaptive importance estimator with multi-resolution selection, allocating higher resolution to query-critical segments. A subset of methods selects frames via text–video and video-context similarity, scoring alignment under budget constraints~\cite{yu2025framevoyagerlearningqueryframes,yao2025generative}.


\noindent \textbf{Video temporal grounding.}
Video Temporal Grounding~\cite{ren2024timechat, wang2024grounded, qian2024momentor, di2024grounded} is a common task in video understanding that associates specific video moments with their corresponding timestamps, while Temporal Localization focuses on accurately identifying these moments within untrimmed videos~\cite{liu2024bench, anne2017localizing, li2024multi}. Current Video-LLMs~\cite{shen2024longvu, wang2024grounded, huang2024lita} have begun to leverage temporal grounding for frame selection by linking video with temporal cues; however, existing methods~\cite{huang2025frag,yu2023self,yu2025frame} mostly focus on single-time retrieval, which take descriptive annotations as input, limiting their generality and robustness in handling diverse real-world scenarios.


In this paper, our VideoITG leverages instructed temporal grounding, automated annotation, and a plug-and-play design to align sampling with user instructions, achieving superior performance and scalability on multimodal video understanding benchmarks.

\section{VideoITG-40K: dataset construction}

\subsection{VidThinker: automated annotation pipeline}
\label{pipeline}
When humans search for information in long videos, they typically proceed in three steps: (i) extracting key cues from the instruction, (ii) retrieving a coarse temporal window, and (iii) fine-grained localization of the target event. We therefore propose \textit{VidThinker}, a fully automated and interpretable pipeline that mimics this three-step reasoning for instruction-guided temporal localization. It comprises three interdependent stages—Instructed Clip Captioning, Instructed Clip Retrieval, and Instructed Frame Localization—that progressively narrow the search space while strengthening alignment with the instruction.

\noindent\textbf{i) Instructed Clip Captioning:} The video $v$ is uniformly divided into short clips (5 seconds each), denoted as $\{v_i\}_{i=0}^n$. For each segment, we employ LLM to extract salient phrases that capture the core information needed to fulfill the instruction. For example, given the question ($q = $``\textit{What does the man playing the drums do with his feet as he plays the drum?}'') and the answer ($a = $``\textit{moves his feet}''), the system distills the essential action phrase: $k =$``\textit{The man playing the drums moves his feet and hits the drums with his hands.}'' We then input the extracted phrases alongside raw video clips into the VLM to generate clip-level descriptions $\{c_i\}_{i=0}^n$ in a recurrent manner. The extracted phrases serve as reference cues to guide the model's attention towards salient elements within each clip.  However, the VLM strictly adheres to visual evidence and it only incorporates information from the extracted phrases when it is explicitly observable in the current clip. This ensures that the system will not hallucinate or infer content solely based on the extracted phrases, maintaining descriptions grounded in visual content. The process can be formulated as follows:
\begin{equation}
    k = \mathrm{LLM}(q, a),\quad c_i = \mathrm{VLM}(k, v_i).
\end{equation}
Conditioning on these instruction- and answer-derived cues, we ensure each segment's annotation is relevant and informative, facilitating precise instructed temporal grounding.

\noindent\textbf{ii) Instructed Clip Retrieval:} The generated clip descriptions $ \{c_i\}_{i=0}^n$ are organized sequentially and evaluated by an LLM for the relevance to the QA pairs. Instead of simply assigning binary relevance scores, the LLM is instructed to perform chain-of-thought reasoning, explicitly considering both keyword matches and temporal relationships to directly output the indexes of relevant clips:
\begin{equation}
    \mathcal{I}_\mathrm{rel-clip} = \mathrm{LLM}(\{c_i\}_{i=0}^n, q, a).
\end{equation}
The chain-of-thought prompting requires the model to justify its selections based on both semantic and temporal cues, rather than relying solely on trivial keyword matching. This automation significantly improves the efficiency and the interpretability of relevant segment selection.
    
\noindent\textbf{iii) Instructed Frame Localization:} After coarse localization of video segment, \textit{VidThinker} further refines the annotation by selecting key frames according to the instruction type. For each frame within the candidate segment, we prompt a LLM to perform a binary classification task: given the QA pair and a single frame, the LLM determines whether the frame is relevant (\texttt{yes}) or not (\texttt{no}) to the instruction.
Formally, for each frame $f_i$ in the candidate segment, the LLM is prompted as follows:
\begin{equation}
    y_i = \mathrm{LLM}(f_i, q, a), \quad \text{where} \quad y_i \in \{\texttt{yes}, \texttt{no}\},
\end{equation}
where $y_i$ indicates whether frame $f_i$ is relevant to the QA.
Only frames with positive responses ($y_i = \texttt{yes}$) are retained as the final temporal grounding results. This instruction-guided filtering allows \textit{VidThinker} to achieve high precision in identifying the most informative frames for instructions.

\subsection{Fine-grained grounding instruction}
\label{taxonomy}
We adopt fine-grained frame selection strategies tailored to each instruction type, ensuring that the visual evidence matches the reasoning needs of each QA task. Since different instructions demand varying visual understanding, we categorize instructions by whether they require static semantics, dynamic motion, both, or no explicit cues at all (video-level). For each type, we adopt a matching frame selection strategy to align visual evidence with QA reasoning needs.

\begin{itemize}[leftmargin=6mm, itemsep=2pt, parsep=0pt, topsep=2pt]
    \item \textbf{Semantic only}: Instructions query static appearance cues (e.g., people, objects, scenes). For example: \textit{"What did the man do before getting into the car?"} \textit{VidThinker} selects frames revealing the man's clothing and the guitar. After relevant segment localization, we select diverse frames that capture representative semantic clues to ensure comprehensive coverage. Concretely, we extract CLIP features per frame and compute cosine similarity between adjacent frames; a frame is retained when its similarity to the last selected keyframe falls below a scene-change threshold. Further algorithmic details are provided in the appendix.

    \item \textbf{Motion only}: Instructions focus on dynamic patterns (e.g., type, speed, direction). We adopt fixed-rate sampling within the localized segment to capture motion progression. For example: \textit{"How does the person jump off the diving board?"} \textit{VidThinker} selects frames spanning takeoff, mid-air, and water entry.
    
    \item \textbf{Semantic \& Motion}: Instructions jointly require static semantics and dynamic changes. We apply fixed-rate sampling in motion-relevant regions while preserving semantically informative frames, balancing both needs. For example: \textit{"Could you describe the camera movement in the video?"} \textit{VidThinker} selects frames showing hand drumming and foot movement simultaneously.
    
    \item \textbf{Non Clues}: Entire video-level instructions without clear semantic or motion anchors. We samle a compact yet diverse set of frames across the entire video (e.g., beginning–middle–end) to ensure holistic coverage with minimal redundancy. For example: \textit{"Please describe the video in detail."}
\end{itemize}

\subsection{Dataset statistics}
\label{statistics}
Leveraging the proposed \textit{VidThinker} pipeline, we construct \textbf{VideoITG-40K} from LLaVA-Video~\cite{zhang2024video}: 40K videos and 500K instruction-grounded annotations for temporal grounding. The entire annotation is automated by \textit{VidThinker}, ensuring efficiency, consistency, and alignment with diverse instruction types. The videos average 120s and cover three duration bands (30–60s, 1–2min, 2–3min). Each video has 10–15 QA pairs (multiple-choice and open-ended). As summarized in Table~\ref{tab:dataset_statistic}, VideoITG-40K is nearly 4$\times$ larger than DiDeMo~\cite{anne2017localizing} (10.6K) and QVHighlights~\cite{lei2021detecting} (10.2K), and far exceeds QuerYD~\cite{oncescu2021queryd} (2.6K) and HiREST~\cite{zala2023hierarchical} (3.4K). Unlike prior descriptive-query datasets, VideoITG-40K is explicitly instruction-guided, enabling precise, query-conditioned temporal localization.

\begin{table}[h]
\renewcommand{\arraystretch}{1.0}
\centering
\caption{
\textbf{Comparison of dataset statistics for temporal grounding and highlight detection datasets.}
}
\vspace{-2mm}
\Large
\resizebox{0.47\textwidth}{!}{
\begin{tabular}{l|c|c|c|c}
\toprule
\textbf{Dataset} & \textbf{\# Videos} & \textbf{\# Queries} & \textbf{Avg. Duration} & \textbf{Instructed?} \\
\midrule
DiDeMo~\cite{anne2017localizing} & 10.6K & 41.2K & 29s & No \\
QuerYD~\cite{oncescu2021queryd} & 2.6K & 32K & \textbf{278s} & No \\
HiREST~\cite{zala2023hierarchical} & 3.4K & 8.6K & 263s & No \\
Charades-STA~\cite{gao2017tall} & 6.7K & 16.1K & 30s & No \\
QVHighlights~\cite{lei2021detecting} & 10.2K & 10.3K & 150s & No \\
\midrule
\rowcolor[HTML]{DAEFF9} VideoITG-40K & \textbf{40K} & \textbf{500K} & 120s & \textbf{Yes} \\
\bottomrule
\end{tabular}
}
\label{tab:dataset_statistic}
\vspace{-2mm}
\end{table}

\begin{figure*}[t]
\vspace{-2mm}
\centering
\includegraphics[width=0.94\linewidth]{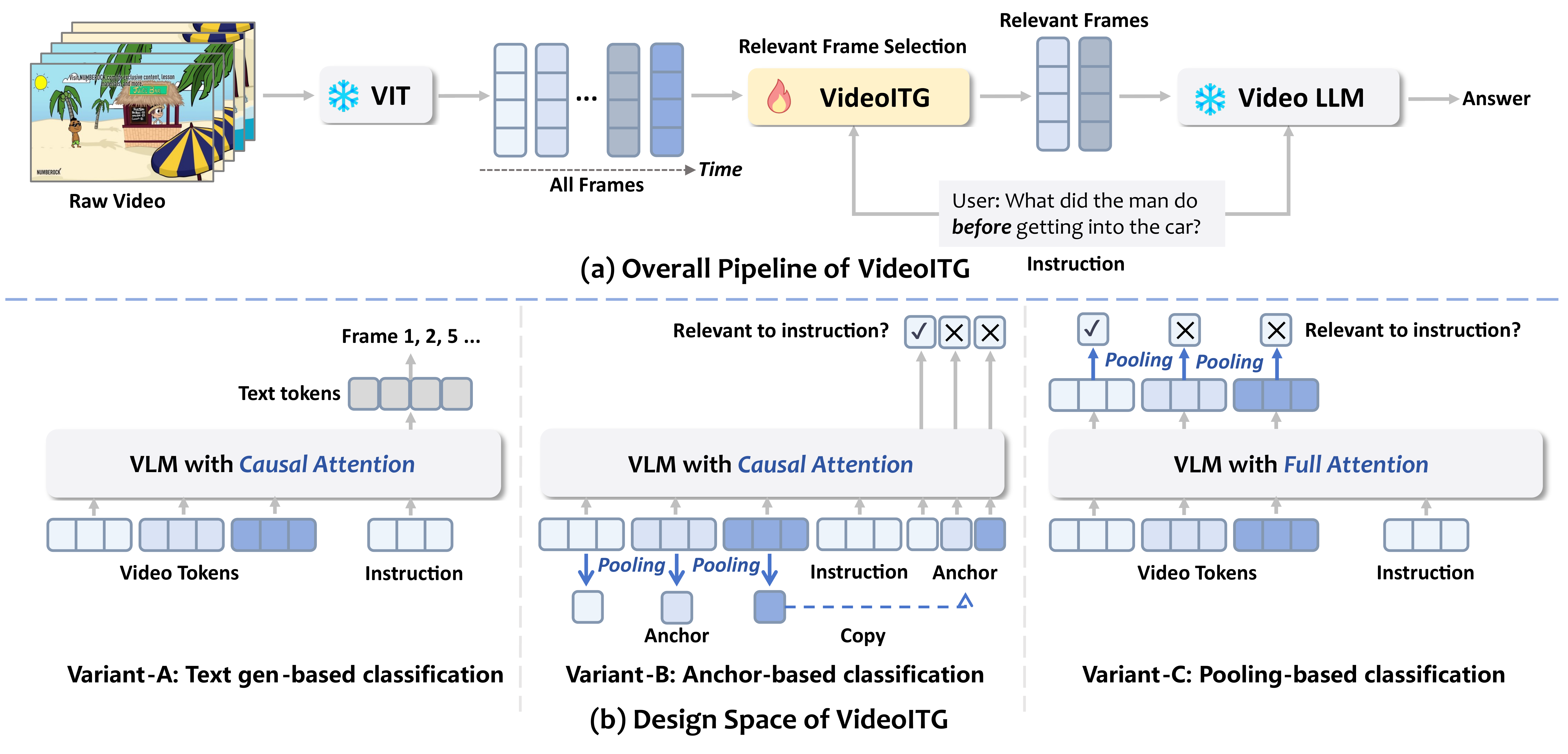}
\vspace{-3mm}
\caption{\textbf{VideoITG model design}: (A) Text generation aligns video and language tokens for sequential predictions. (B) Classification with causal attention utilizes anchor tokens for temporal cue management. (C) Classification with full attention facilitates interaction across visual and text tokens without anchors.}
\label{fig:model}
\vspace{-2mm}
\end{figure*}

\section{VideoITG: model design}

In this section, we explore how to utilize our VideoITG-40K dataset to train the model for the \textbf{Instructed Temporal Grounding} task, aiming to optimize video frame selection and enhance the performance of Video-LLMs. As illustrated in Fig.~\ref{fig:model}, our framework comprises three modules: 
(1) a vision encoder (\textit{e.g.}, ViT) that maps video frames into text-aligned visual features $F$, 
(2) a VideoITG module that performs instruction-guided frame selection $\mathcal{I}_\mathrm{rel}$, 
and (3) a VideoLLM that generates answers $a$ conditioned on the selected frames $F_{\mathcal{I}_\mathrm{rel}}$ and the question $q$. The process can be described as folllows:
\begin{align}
    F &= \mathrm{VIT}(v) \\
    \mathcal{I}_\mathrm{rel} &= \mathrm{VideoITG}(F, q) \\
    a &= \mathrm{VideoLLM}(F_{ \mathcal{I}_\mathrm{rel}}, q)
\end{align}
The VideoITG module follows a plug-and-play design philosophy, driven by two core objectives: 
(1) enhancing the alignment between visual and language tokens to improve instruction following, 
and (2) strengthening contextual encoding to capture multi-granular temporal cues. With the above considerations, we develop three model variants: text generation-based classification, anchor-based classification, and pooling-based classification, as illustrated in Fig.~\ref{fig:model} (b).

\noindent\textbf{Variant A: Text-generation-based classification.}
As shown in Fig.~\ref{fig:model}(b, left), this variant reformulates the Instructed Temporal Grounding task as a next-token prediction problem, where the model sequentially outputs text tokens conditioned on video and instruction features. 
This formulation naturally aligns with the core training paradigm of existing Video-LLMs, thereby preserving their strong vision-language alignment and instruction-following abilities. 
Similar generative frameworks have also been adopted in prior time-sensitive models such as TimeChat~\cite{ren2024timechat} and Grounded-VideoLLM~\cite{wang2024grounded}.
\begin{table*}[t]
\small
\centering
\caption{\textbf{Results with different selection methods.} We bold the best results on each benchmark under the same Answering LMM. When comparing with Q-Frame, we adopt a slow–fast strategy of 4 high-resolution, 8 medium-resolution, and 32 low-resolution frames, where the total number of tokens is equivalent to that of 8 high-resolution frames.}
\vspace{-2mm}
\resizebox{\textwidth}{!}{%
\begin{tabular}{ll|c|c|c|ccc|c}
\toprule
\multirow{2}{*}{\textbf{Selection Methods}} &
\multirow{2}{*}{\textbf{Answering LMM}} &
\multirow{2}{*}{\textbf{Frames}} &
\multicolumn{1}{c|}{\textbf{LongVideoBench}} &
\multicolumn{1}{c|}{\textbf{MLVU}} &
\multicolumn{3}{c|}{\textbf{VideoMME}} &
\multirow{2}{*}{\textbf{Average}} \\
\cmidrule(lr){4-4} \cmidrule(lr){5-5} \cmidrule(lr){6-8}
& & & \textit{8 min} & \textit{12 min} & \textit{S (2 min)} & \textit{M (10 min)} & \textit{L (40 min)} & \\
\midrule
Uniform & LLaVA-OneVision-7B & 8 & 54.2 & 58.9 & 63.6 & 52.0 & 45.7  & 54.9 \\
BOLT~\cite{liu2025boltboostlargevisionlanguage} & LLaVA-OneVision-7B & 8 & 56.1 & 63.4 & 66.8   & 54.2 & 47.3 & 57.6 \\
Frame-VOYAGER~\cite{yu2025framevoyagerlearningqueryframes} & LLaVA-OneVision-7B & 8 & -    & 65.6 & 67.3 & 56.3 & 48.9   & 59.5 \\
\rowcolor[HTML]{DAEFF9}VideoITG-8B & LLaVA-OneVision-7B & 8 & \textbf{60.1} & \textbf{68.7} & \textbf{72.0} & \textbf{57.7} & \textbf{49.4} & \textbf{61.6} \\
\midrule
Uniform & Qwen2-VL & 8 & 53.5 & 56.9 & 65.0 & 50.7 & 45.3 & 54.3 \\
Q-Frame~\cite{zhang2025q} & Qwen2-VL & 8+16+32 & 58.4 & 65.4 & 69.4 & 57.1 & 48.3 & 59.7 \\
\rowcolor[HTML]{DAEFF9}VideoITG-8B & Qwen2-VL & 8+16+32 & \textbf{58.6} & \textbf{66.6} & \textbf{69.8} & \textbf{57.3} & \textbf{49.2} & \textbf{60.3} \\
\midrule
Uniform & LLaVA-Video-7B & 64 & 59.9 & 70.2 & 75.8 & 63.0 & 54.7 & 64.7 \\
AKS~\cite{tang2025adaptive} & LLaVA-Video-7B & 32 & 59.6 & 74.3 & 75.1 & 63.9 & 51.7 & 64.9 \\
QuoTA~\cite{luo2025quota} & LLaVA-Video-7B & 64 & 59.0 & 71.9 & 71.1 & 58.8 & 52.2 & 62.6 \\
Gen-S~\cite{yao2025generative} & LLaVA-Video-7B & 54/50 & 63.3 & 73.4 & - & - & - & - \\
\rowcolor[HTML]{DAEFF9}VideoITG-8B & LLaVA-Video-7B & 32 & 61.6 & 74.6 & \textbf{77.3} & 65.9 & 55.2 & 66.9 \\
\rowcolor[HTML]{DAEFF9}VideoITG-8B & LLaVA-Video-7B & 64 & 60.9 & \textbf{76.3} & 76.1 & \textbf{66.0} & \textbf{56.1} & \textbf{67.1} \\
\bottomrule
\end{tabular}}
\label{tab:main_comparisons}
\vspace{-2mm}
\end{table*}

\noindent\textbf{Variant B: Anchor-based classification.}
To move beyond token-by-token generation, this variant adopts a discriminative paradigm that directly classifies visual tokens at the frame level (Fig.~\ref{fig:model}(b, middle)). 
We initialize the model from a pretrained Video-LLM while maintaining its causal attention mask to retain temporal consistency. 
However, the causal mask prevents visual tokens from accessing the instruction beforehand and restricts early frames from leveraging subsequent temporal cues. 
To mitigate this limitation, we insert an \emph{anchor token} after the instruction, serving as a temporal mediator for each frame.  
Formally, for a video frame at timestamp $t$, the anchor token $A^t$ is derived by global averaging over all spatial locations:
\begin{equation}
A^t = \frac{1}{M}\sum\nolimits_{i,j} F_{ij}^t, \quad t \in [1, T],
\end{equation}
where $F_{ij}^t$ denotes the visual feature at grid $(i,j)$ of the $t$-th frame and $M$ is the total number of patches per frame. 
The set $\{A^t\}_{t=1}^T$ bridges temporal dependencies across frames under causal attention.

\noindent\textbf{Variant C: Pooling-based classification.}
As the causal attention mask restricts inter-frame communication, we further remove this constraint to enable full bidirectional attention between visual and textual tokens (Fig.~\ref{fig:model}(b, right)). 
For each frame, we aggregate its visual tokens through average pooling, followed by a classification head that determines instruction relevance, without introducing explicit anchor tokens.
This full-attention design enriches temporal context modeling across frames and facilitates stronger interaction between instructions and visual evidence.

\section{Experiments}

\begin{table*}[h]
\small
\centering
\caption{\textbf{Performance comparison of VideoITG integrated with different Video-LLMs, varying in both the size of the answering LLM and the number of sampled frames.}``UNI-$k$'' denotes UNIform sampling of $k$ frames, while ``ITG-$k$'' refers to selecting the Top $k$ frames based on relevance scores generated by our proposed VideoITG.
}
\vspace{-2mm}
\resizebox{\textwidth}{!}{
\begin{tabular}{ll|c|c|ccc|c|c}
\toprule
\multirow{2}{*}{\textbf{LMM}} & \multirow{2}{*}{\textbf{Selection}} & \multicolumn{1}{c|}{\textbf{LongVideoBench}} & \multicolumn{1}{c|}{\textbf{MLVU}} & \multicolumn{3}{c|}{\textbf{VideoMME}} & \multicolumn{1}{c|}{\textbf{CG-Bench-mini}} & \multirow{2}{*}{\textbf{Average}} \\
\cmidrule(lr){3-3} \cmidrule(lr){4-4} \cmidrule(lr){5-7} \cmidrule(lr){8-8} 
& & \textit{8min} & \textit{12min} & \textit{S (2 min)} & \textit{M (10 min)} & \textit{L (40 min)} & \textit{27min}  \\
\midrule
\multirow{2}{*}{InternVL2.5-8B} & UNI-32 & 58.3 & 66.4 & 75.1 & 61.7 & 53.1 & 37.7 & 58.7 \\
& ITG-32  & \cellcolor[HTML]{DAEFF9} \textbf{61.9} \textbf{(+3.6)} & \cellcolor[HTML]{DAEFF9} \textbf{75.0} \textbf{(+8.6)} & \cellcolor[HTML]{DAEFF9} \textbf{78.0} \textbf{(+2.9)} & \cellcolor[HTML]{DAEFF9} \textbf{67.1} \textbf{(+5.4)} & \cellcolor[HTML]{DAEFF9} \textbf{56.9} \textbf{(+3.8)} & \cellcolor[HTML]{DAEFF9} \textbf{46.7} \textbf{(+9.0)} & \cellcolor[HTML]{DAEFF9} \textbf{64.3} \textbf{(+5.6)} \\
\midrule
\multirow{2}{*}{InternVL2.5-26B} & UNI-32 & 55.6 & 71.3 & 78.1 & 67.1 & 56.9 & 40.6 & 61.6 \\
& ITG-32  & \cellcolor[HTML]{DAEFF9} \textbf{63.0} \textbf{(+7.4)} & \cellcolor[HTML]{DAEFF9} \textbf{78.9} \textbf{(+7.6)} & \cellcolor[HTML]{DAEFF9} \textbf{80.8} \textbf{(+2.7)} & \cellcolor[HTML]{DAEFF9} \textbf{69.0} \textbf{(+1.9)} & \cellcolor[HTML]{DAEFF9} \textbf{59.9} \textbf{(+3.0)} & \cellcolor[HTML]{DAEFF9} \textbf{48.7} \textbf{(+8.1)} & \cellcolor[HTML]{DAEFF9} \textbf{66.7} \textbf{(+5.1)} \\
\midrule
\multirow{2}{*}{InternVL3.5-8B} & UNI-32 & 60.0 & 70.0 & 77.0 & 62.4 & 53.4 & 40.9 & 60.6 \\
& ITG-32  & \cellcolor[HTML]{DAEFF9} \textbf{65.7} \textbf{(+5.7)} & \cellcolor[HTML]{DAEFF9} \textbf{74.1} \textbf{(+4.1)} & \cellcolor[HTML]{DAEFF9} \textbf{78.4} \textbf{(+1.4)} & \cellcolor[HTML]{DAEFF9} \textbf{65.9} \textbf{(+3.5)} & \cellcolor[HTML]{DAEFF9} \textbf{59.0} \textbf{(+5.6)} & \cellcolor[HTML]{DAEFF9} \textbf{47.6} \textbf{(+6.7)} & \cellcolor[HTML]{DAEFF9} \textbf{65.1} \textbf{(+4.5)} \\
\midrule
\multirow{2}{*}{Qwen3-VL} & UNI-32 & 59.1 & 64.1 & 76.0 & 60.9 & 55.1 & 40.1 & 59.2 \\
& ITG-32  & \cellcolor[HTML]{DAEFF9} \textbf{63.6} \textbf{(+4.5)} & \cellcolor[HTML]{DAEFF9} \textbf{77.2} \textbf{(+13.1)} & \cellcolor[HTML]{DAEFF9} \textbf{79.9} \textbf{(+3.9)} & \cellcolor[HTML]{DAEFF9} \textbf{66.6} \textbf{(+5.7)} & \cellcolor[HTML]{DAEFF9} \textbf{60.3} \textbf{(+5.2)} & \cellcolor[HTML]{DAEFF9} \textbf{47.3} \textbf{(+7.2)} & \cellcolor[HTML]{DAEFF9} \textbf{65.8} \textbf{(+6.6)} \\
\midrule
\multirow{2}{*}{LLaVA-Video-7B} & UNI-32 & 58.7 & 66.8 & 76.3 & 60.3 & 52.7 & 35.8 & 58.4 \\
& ITG-32  & \cellcolor[HTML]{DAEFF9} \textbf{61.6} \textbf{(+2.9)} & \cellcolor[HTML]{DAEFF9} \textbf{74.6} \textbf{(+7.8)} & \cellcolor[HTML]{DAEFF9} \textbf{77.3} \textbf{(+1.0)} & \cellcolor[HTML]{DAEFF9} \textbf{65.9} \textbf{(+5.6)} & \cellcolor[HTML]{DAEFF9} \textbf{55.2} \textbf{(+2.5)} & \cellcolor[HTML]{DAEFF9} \textbf{42.8} \textbf{(+7.0)} & \cellcolor[HTML]{DAEFF9} \textbf{62.9} \textbf{(+4.5)} \\
\midrule
\multirow{2}{*}{Eagle2.5-8B} & UNI-32 & 63.0 & 67.8 & 78.8 & 64.1 & 55.9 & 41.2 & 61.8 \\
& ITG-32  & \cellcolor[HTML]{DAEFF9} \textbf{66.8 (+3.8)} & \cellcolor[HTML]{DAEFF9} \textbf{76.5 (+8.7)} & \cellcolor[HTML]{DAEFF9} \textbf{80.0 (+1.2)} & \cellcolor[HTML]{DAEFF9} \textbf{67.8 (+3.7)} & \cellcolor[HTML]{DAEFF9} \textbf{60.3 (+4.4)} & \cellcolor[HTML]{DAEFF9} \textbf{49.0 (+7.8)} & 
\cellcolor[HTML]{DAEFF9} \textbf{66.7 (+4.9)} \\
\bottomrule
\end{tabular}}
\label{tab:video}
\vspace{-2mm}
\end{table*}



\begin{table*}[!h]
\centering
\caption{\textbf{Empirical studies on the VideoITG-40k dataset and VideoITG model design.} We adopt Variant-C for subsequent experiments. ``No Images" and ``No Videos" indicate that image-text data (LAION-CC-SBU-558K \& LLaVA-OV-SI) or video data (LLaVA-Video-178K) are excluded from pre-training, respectively.}
\vspace{-2mm}
\resizebox{\textwidth}{!}{%
\setlength{\tabcolsep}{8pt}
\begin{tabular}{l|c|ccc|c|c|c}
\toprule
\multirow{2}{*}{\textbf{Abaltion}} &
\multirow{2}{*}{\textbf{Experiment}} &
\multicolumn{3}{c|}{\textbf{Videomme}} &
\multicolumn{1}{c|}{\textbf{MLVU}} &
\multicolumn{1}{c|}{\textbf{LongVideoBench}} &
\multicolumn{1}{c}{\textbf{Average}} \\

&& \textbf{Short} (\%) $\uparrow$ & \textbf{Medium} (\%) $\uparrow$ & \textbf{Long} (\%) $\uparrow$ & \textbf(\%) $\uparrow$ & \textbf(\%) $\uparrow$\\
\midrule

& Variant-A-7B & 51.0 & 44.8 & 44.4 & 45.7 & 56.8 & 48.5 \\
\textbf{Architecture}
& Variant-B-7B & 77.9 & 66.0 & 56.2 & 74.6 & 61.3 & 67.2 \\
\textbf{Design} & \cellcolor[HTML]{DAEFF9}Variant-C-7B &\cellcolor[HTML]{DAEFF9}\textbf{78.0} &\cellcolor[HTML]{DAEFF9}\textbf{67.1} & \cellcolor[HTML]{DAEFF9}56.9 & \cellcolor[HTML]{DAEFF9}\textbf{75.0} & \cellcolor[HTML]{DAEFF9}\textbf{61.9} & \cellcolor[HTML]{DAEFF9}\textbf{67.8} \\
& Variant-C-3B & 77.1 & 64.8 & 56.0 & 74.5 & 61.5 & 66.8\\
\midrule 
\textbf{Dataset}& No Clip Captioning & 77.5 & 63.1 & 53.4 & 73.2 & 61.7 &  65.8  \\
\textbf{Construction} & No Frame Localization & 77.6 & 65.8  & 56.8 & 74.1 & 61.5 & 67.2 \\
\midrule 
\textbf{Pre-training}& No Videos & 77.2 & 64.9 & \textbf{57.4} & 74.5 & 61.6 & 67.1 \\
\textbf{Data} & No Images \& Videos & 76.6 & 63.0  & 54.4 & 69.1 & 58.6 & 64.3 \\
\bottomrule
\end{tabular}}
\vspace{-2mm}
\label{ablation_table_models}
\end{table*}

\subsection{Implementation details}
We follow the training approach of LLaVA-Video~\cite{zhang2024video}, using the pretrained model as the initialization for our VideoITG model's pre-training. We employ SigLIP~\cite{zhai2023sigmoid} as the vision encoder and Qwen2~\cite{wang2024qwen2} as the language model. Initially, we train the MLP projector on image caption datasets with a batch size of 256 and a learning rate of \(1 \times 10^{-3}\). Then, we fine-tune all model parameters on the LLaVA-OV-SI~\cite{li2024llavaonevision} and LLaVA-Video datasets. During this stage, the video frame sampling rate is set to 64, and the LLM's maximum sequence length is set to 16K. We then train the VideoITG model on the proposed VideoITG-40K dataset, adjusting the video sampling rate to 1 fps. 

Throughout training and inference, we employ a dynamic token spatial size strategy~\cite{liu2025oryx}. Across all stages, the LLM's learning rate is \(2 \times 10^{-5}\), and in the final stage, the learning rate for the classification head is \(2 \times 10^{-4}\). To fairly compare with other leading video LMMs, we primarily use results from their original papers. When results are unavailable, we integrate the models into LMMs-Eval~\cite{zhang2024lmms} and assess them under consistent settings. Due to context length constraints, we support up to 512 video frames as input (with 16 visual tokens per frame) for the VideoITG model, from which we select the top 32 frames based on their scores by default. 

\subsection{Main results}

\noindent \textbf{Comparisons with other frame selection methods.}
As shown in Table~\ref{tab:main_comparisons}, our proposed VideoITG demonstrates \textbf{\textit{three notable advantages}} over existing selection strategies: 

\textbf{(i) Consistant Improvement}: VideoITG consistently outperforms Uniform sampling across all benchmarks and model settings. It achieves an average gain of \textbf{+6.7} (54.9$\rightarrow$61.6) on LLaVA-OneVision-7B and \textbf{+6.2} (54.1$\rightarrow$60.3) on Qwen2-VL, with similar trends on LLaVA-Video-7B (+2$\sim$3 points). These steady improvements across diverse models and datasets highlight the \textbf{robustness} and \textbf{general applicability} of our frame selection strategy in enhancing video understanding under challenging scenarios.

\textbf{(ii) Precision in Selection}: When using LLaVA-Video as baseline, VideoITG achieves superior accuracy using only 32 selected frames, outperforming other methods that use \textbf{more frames} (\textit{e.g.}, 50-64) on MLVU and VideoMME, highlighting the effectiveness of our strategy in identifying the most informative frames. As shown in Figure~\ref{fig:frames}, our VideoITG method achieves comparable performance to uniform sampling with 64 frames using only 16 frames on VideoMME, and also outperforms other methods with 32 frames, highlighting the precision of our frame selection strategy. 

\textbf{(iii) Robust Performance:} Compared to training-based methods such as Q-Frame and Frame-Voyager, VideoITG achieves more substantial improvements across three long video understanding benchmarks. For instance, it boosts the average score from 59.5 (Frame-Voyager) and 57.6 (Q-Frame) to \textbf{61.6} and \textbf{60.3} under the same settings. These consistent gains highlight the strong adaptability of VideoITG across diverse scenarios.

We find that on LongVideoBench, increasing frames has limited impact, likely because many tasks are text referring. Our VideoITG also samples at lower resolution, indicating room for finer-grained content recognition.
 
\vspace{1mm}
\noindent\textbf{Results on newer backbone (Qwen3-VL).}
To verify that our gains are not tied to a specific backbone generation, we evaluate VideoITG on the newer Qwen3-VL model. The results are included in Table~\ref{tab:video}, showing consistent improvements over the UNI-32 baseline across LongVideoBench, MLVU, VideoMME (S/M/L), and CG-Bench-mini.

\noindent\textbf{Extension to more Video-LLMs.} Extending our VideoITG on diverse Video-LLMs with difeerent model scales further presents its \textbf{\textit{two compiling advantages}}: 

\textbf{(i) Model Size Scalability}:
Table~\ref{tab:video} demonstrates that integrating VideoITG with Video-LLMs of different sizes consistently yields substantial performance improvements over uniform sampling. For example, on InternVL2.5-8B, VideoITG improves the average score from $58.7\%$ to $64.3\%$ ($+5.6$), while on the larger InternVL2.5-26B, the improvement is from $61.6\%$ to $66.7\%$ ($+5.1$). Notably, InternVL2.5-8B with VideoITG even surpasses the InternVL2.5-26B baseline on both average score ($64.3\%$ vs. $61.6\%$) and long-video benchmarks such as CG-Bench ($46.7\%$ vs. $40.6\%$), indicating that effective frame selection can provide greater gains than simply increasing model size. 

\begin{figure*}[!h]
\centering
\includegraphics[width=\linewidth]{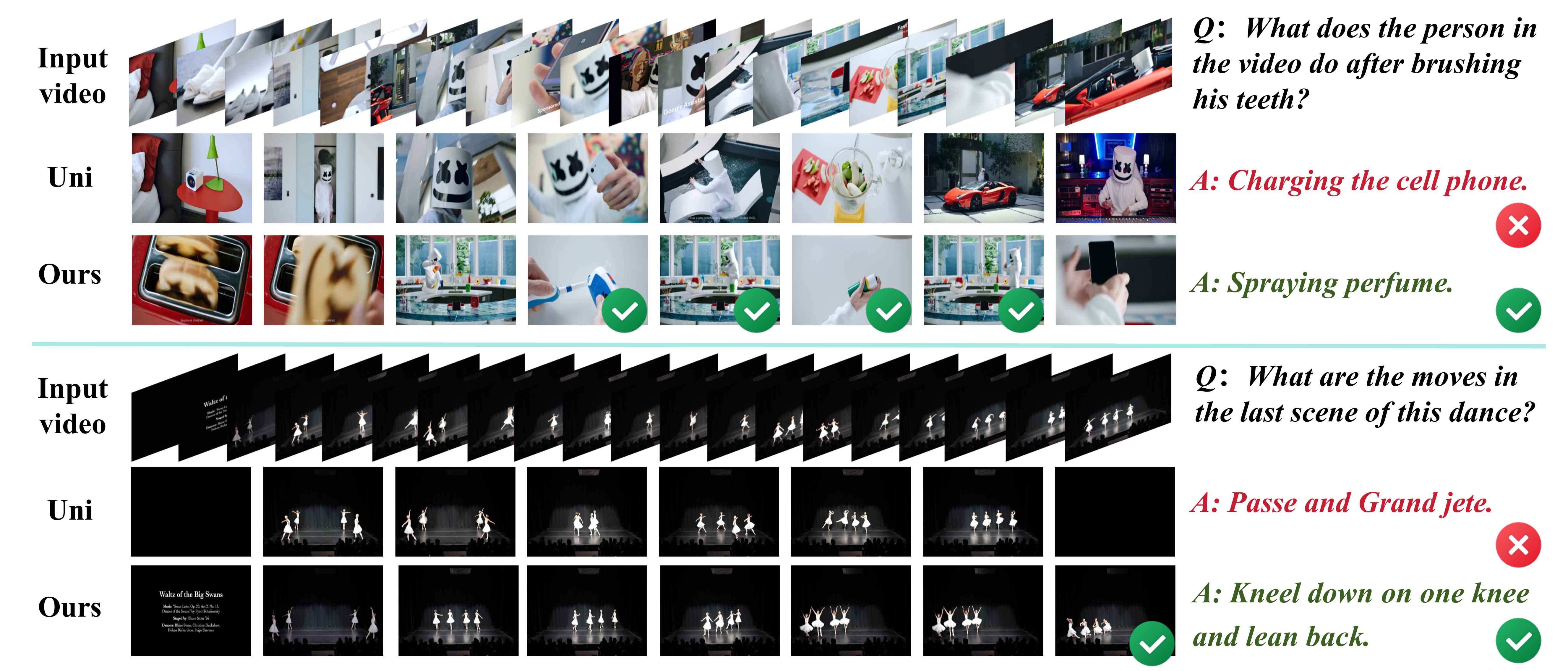}
\vspace{-7mm}
\caption{\textbf{Two examples of how different sampling strategies impact video understanding.} We mark the identified key frames that directly answer the question with green check-marks.}
\label{vis}
\end{figure*}

(ii) \textbf{Model Diversity Adaptability}:
We further evaluate VideoITG across diverse Video-LLMs, each trained on different data distributions and objectives. The results show that VideoITG consistently outperforms uniform sampling for all models and benchmarks. For instance, on LLaVA-Video-7B, VideoITG raises the average score by $4.5$ \% ($62.9\%$ vs. $58.4\%$), and on Eagle2.5-8B, the improvement reaches $4.9$ points ($66.7\%$ vs. $61.8\%$). These consistent gains across models with varying architectures and training data highlight the strong adaptability and robustness of our method.

\subsection{Ablation on VideoITG design choices}
Table~\ref{ablation_table_models} presents a comprehensive analysis on the design of VideoITG, directly supporting our key contributions.

\noindent\textbf{Architecture Design.} 
First, we compare the three variants of our model architecture in Fig.~\ref{fig:model}. We observe that Variant A, which is based on the text generation paradigm, performs the worst. One possible reason is that text generation models trained with the next-token prediction paradigm suffer from sparse supervision due to teacher forcing, where previous frame selections influence subsequent ones, making the training process less efficient compared to discriminative classification models. We find that Variant C with full-attention outperforms Variant B with causal attention. This improvement may be attributed to full-attention's larger receptive field, which enables global temporal relationship modeling and allows all tokens to access the textual query. Moreover, model scale yields a consistent but modest gain: the 7B model surpasses the 3B counterpart across all benchmarks, raising the overall average from 66.8\% to 67.8\%.

\noindent\textbf{Dataset Construction.} We analyze our data annotation strategies to demonstrate the effectiveness of our pipeline. Ablation studies show that the performance degrades when Instructed Clip Captioning are removed (using a VLM to directly select temporal boundaries based on visual inputs), with accuracy dropping from 56.9\% to 53.4\% on Videomme Long videos and from 75.0\% to 73.2\% on MLVU. This demonstrates that ensuring information diversity is crucial for maintaining comprehensive feature  representation of videos. Similarly, removing Instructed Frame Localization (learning all frame index within corase temporal boundaries using VideoITG Model) decreases performance, particularly on Videomme Medium videos (from 67.1\% to 65.8\%). These results confirm that both stages are essential for optimal model performance and validate our data construction approach of the VideoITG-40K dataset.

\noindent\textbf{Pre-training Data.} Finally, we investigate the impact of vision-language alignment pre-training on model performance. Our experiments reveal that removing video pre-training causes modest performance changes across benchmarks. This suggests that the benefits of video data for instructed temporal grounding tasks primarily stem from effective visual context length, yet this impact is relatively minor compared to vision-language alignment. This observation is further validated if we eliminate both image and video pre-training data, starting from a text-only large language model, where performance drops dramatically, with accuracy decreasing from 75.0\% to 69.1\% on MLVU and from 61.9\% to 58.6\% on LongVideoBench. This substantial degradation underscores that robust vision-language alignment is crucial to effective VideoITG training.

\subsection{Visualization}
In Fig.~\ref{vis}, we compare uniform sampling and VideoITG sampling of 8 frames from the VideoMME~\cite{fu2024videomme} Benchmark. In the first case, VideoITG captures both brushing teeth and spraying perfume actions, enabling correct temporal ordering, while uniform sampling misses key cues. In the second case, VideoITG accurately captures rapid consecutive movements at the end, whereas uniform sampling fails to do so, leading to incomplete video understanding.

\section{Conclusion}
In this paper, we presented VideoITG, a novel framework for instruction-aligned frame selection in Video-LLMs. The key to our approach was the \textit{VidThinker} pipeline, which mimics human annotation by generating detailed, instruction-guided clip descriptions, retrieving relevant segments, and performing fine-grained frame selection. Using this pipeline, we constructed the VideoITG-40K dataset with 40K videos and 500K temporal grounding annotations. Based on this resource, we developed plug-and-play VideoITG models that leverage visual-language alignment and reasoning to handle diverse temporal grounding tasks. Experiments showed that VideoITG consistently improves Video-LLMs' performance across multiple video understanding benchmarks, highlighting its effectiveness and potential for advancing instruction-driven video understanding.


{
    \small
    \bibliographystyle{ieeenat_fullname}
    \bibliography{main}

\begin{thebibliography}{74}
\providecommand{\natexlab}[1]{#1}
\providecommand{\url}[1]{\texttt{#1}}
\expandafter\ifx\csname urlstyle\endcsname\relax
  \providecommand{\doi}[1]{doi: #1}\else
  \providecommand{\doi}{doi: \begingroup \urlstyle{rm}\Url}\fi

\bibitem[Anne~Hendricks et~al.(2017)Anne~Hendricks, Wang, Shechtman, Sivic, Darrell, and Russell]{anne2017localizing}
Lisa Anne~Hendricks, Oliver Wang, Eli Shechtman, Josef Sivic, Trevor Darrell, and Bryan Russell.
\newblock Localizing moments in video with natural language.
\newblock In \emph{ICCV}, 2017.

\bibitem[Bai et~al.(2025)Bai, Cai, Chen, Chen, Chen, Cheng, Deng, Ding, Gao, Ge, et~al.]{bai2025qwen3}
Shuai Bai, Yuxuan Cai, Ruizhe Chen, Keqin Chen, Xionghui Chen, Zesen Cheng, Lianghao Deng, Wei Ding, Chang Gao, Chunjiang Ge, et~al.
\newblock Qwen3-vl technical report.
\newblock \emph{arXiv preprint arXiv:2511.21631}, 2025.

\bibitem[Brohan et~al.(2023)Brohan, Brown, Carbajal, Chebotar, Chen, Choromanski, Ding, Driess, Dubey, Finn, et~al.]{brohan2023rt}
Anthony Brohan, Noah Brown, Justice Carbajal, Yevgen Chebotar, Xi Chen, Krzysztof Choromanski, Tianli Ding, Danny Driess, Avinava Dubey, Chelsea Finn, et~al.
\newblock {RT-2}: Vision-language-action models transfer web knowledge to robotic control.
\newblock \emph{arXiv:2307.15818}, 2023.

\bibitem[Chai et~al.(2025)Chai, Song, Du, Meng, Madhavan, Bar-Tal, Hwang, Xie, and Manning]{chai2025auroracap}
Wenhao Chai, Enxin Song, Yilun Du, Chenlin Meng, Vashisht Madhavan, Omer Bar-Tal, Jeng-Neng Hwang, Saining Xie, and Christopher~D Manning.
\newblock {AuroraCap}: Efficient, performant video detailed captioning and a new benchmark.
\newblock 2025.

\bibitem[Chen et~al.(2024{\natexlab{a}})Chen, Liu, Huang, He, Pei, Xu, Wang, Lu, and Wang]{chen2024cg}
Guo Chen, Yicheng Liu, Yifei Huang, Yuping He, Baoqi Pei, Jilan Xu, Yali Wang, Tong Lu, and Limin Wang.
\newblock {CG-Bench}: Clue-grounded question answering benchmark for long video understanding.
\newblock \emph{arXiv:2412.12075}, 2024{\natexlab{a}}.

\bibitem[Chen et~al.(2025{\natexlab{a}})Chen, Li, Wang, Jiang, Liu, Lu, Huang, Byeon, Le, Rintamaki, et~al.]{chen2025eagle}
Guo Chen, Zhiqi Li, Shihao Wang, Jindong Jiang, Yicheng Liu, Lidong Lu, De-An Huang, Wonmin Byeon, Matthieu Le, Tuomas Rintamaki, et~al.
\newblock Eagle 2.5: Boosting long-context post-training for frontier vision-language models.
\newblock \emph{arXiv preprint arXiv:2504.15271}, 2025{\natexlab{a}}.

\bibitem[Chen et~al.(2024{\natexlab{b}})Chen, Lv, Wu, Lin, Song, Gao, Liu, Gao, Mao, and Shou]{chen2024videollm}
Joya Chen, Zhaoyang Lv, Shiwei Wu, Kevin~Qinghong Lin, Chenan Song, Difei Gao, Jia-Wei Liu, Ziteng Gao, Dongxing Mao, and Mike~Zheng Shou.
\newblock {VideoLLM-online}: Online video large language model for streaming video.
\newblock In \emph{CVPR}, 2024{\natexlab{b}}.

\bibitem[Chen et~al.(2024{\natexlab{c}})Chen, Wei, Li, Dong, Zhang, Zang, Chen, Duan, Lin, Tang, et~al.]{chensharegpt4video}
Lin Chen, Xilin Wei, Jinsong Li, Xiaoyi Dong, Pan Zhang, Yuhang Zang, Zehui Chen, Haodong Duan, Bin Lin, Zhenyu Tang, et~al.
\newblock {ShareGPT4Video}: Improving video understanding and generation with better captions.
\newblock \emph{arXiv:2406.04325}, 2024{\natexlab{c}}.

\bibitem[Chen et~al.(2025{\natexlab{b}})Chen, Di, and Xie]{chen2025grounded}
Qirui Chen, Shangzhe Di, and Weidi Xie.
\newblock Grounded multi-hop videoqa in long-form egocentric videos.
\newblock In \emph{AAAI}, 2025{\natexlab{b}}.

\bibitem[Chen et~al.(2024{\natexlab{d}})Chen, Guo, and Wang]{chen2024large}
Yang Chen, Sheng Guo, and Limin Wang.
\newblock A large-scale study on video action dataset condensation.
\newblock \emph{arXiv:2412.21197}, 2024{\natexlab{d}}.

\bibitem[Cheng et~al.(2024)Cheng, Leng, Zhang, Xin, Li, Chen, Zhu, Zhang, Luo, Zhao, et~al.]{cheng2024videollama}
Zesen Cheng, Sicong Leng, Hang Zhang, Yifei Xin, Xin Li, Guanzheng Chen, Yongxin Zhu, Wenqi Zhang, Ziyang Luo, Deli Zhao, et~al.
\newblock {VideoLLaMA 2}: Advancing spatial-temporal modeling and audio understanding in video-llms.
\newblock \emph{arXiv:2406.07476}, 2024.

\bibitem[Dao(2024)]{dao2023flashattention2}
Tri Dao.
\newblock Flash{A}ttention-2: Faster attention with better parallelism and work partitioning.
\newblock In \emph{ICLR}, 2024.

\bibitem[Dao et~al.(2022)Dao, Fu, Ermon, Rudra, and R{\'e}]{dao2022flashattention}
Tri Dao, Daniel~Y. Fu, Stefano Ermon, Atri Rudra, and Christopher R{\'e}.
\newblock {FlashAttention}: Fast and memory-efficient exact attention with {IO}-awareness.
\newblock In \emph{NeurIPS}, 2022.

\bibitem[Di and Xie(2024)]{di2024grounded}
Shangzhe Di and Weidi Xie.
\newblock Grounded question-answering in long egocentric videos.
\newblock In \emph{CVPR}, 2024.

\bibitem[Fu et~al.(2024{\natexlab{a}})Fu, Dai, Luo, Li, Ren, Zhang, Wang, Zhou, Shen, Zhang, et~al.]{fu2024videomme}
Chaoyou Fu, Yuhan Dai, Yondong Luo, Lei Li, Shuhuai Ren, Renrui Zhang, Zihan Wang, Chenyu Zhou, Yunhang Shen, Mengdan Zhang, et~al.
\newblock {Video-MME}: The first-ever comprehensive evaluation benchmark of multi-modal llms in video analysis.
\newblock \emph{arXiv:2405.21075}, 2024{\natexlab{a}}.

\bibitem[Fu et~al.(2024{\natexlab{b}})Fu, Lin, Long, Shen, Zhao, Zhang, Dong, Wang, Yin, Ma, et~al.]{fu2024vita}
Chaoyou Fu, Haojia Lin, Zuwei Long, Yunhang Shen, Meng Zhao, Yifan Zhang, Shaoqi Dong, Xiong Wang, Di Yin, Long Ma, et~al.
\newblock {VITA}: Towards open-source interactive omni multimodal llm.
\newblock \emph{arXiv:2408.05211}, 2024{\natexlab{b}}.

\bibitem[Gao et~al.(2017)Gao, Sun, Yang, and Nevatia]{gao2017tall}
Jiyang Gao, Chen Sun, Zhenheng Yang, and Ram Nevatia.
\newblock {TALL}: Temporal activity localization via language query.
\newblock In \emph{ICCV}, 2017.

\bibitem[Han et~al.(2025)Han, Huang, Shi, Zhuo, Su, Zhang, Zhou, Qi, Liao, and Liu]{han2025videoespresso}
Songhao Han, Wei Huang, Hairong Shi, Le Zhuo, Xiu Su, Shifeng Zhang, Xu Zhou, Xiaojuan Qi, Yue Liao, and Si Liu.
\newblock Videoespresso: A large-scale chain-of-thought dataset for fine-grained video reasoning via core frame selection.
\newblock In \emph{CVPR}, 2025.

\bibitem[Huang et~al.(2024)Huang, Liao, Radhakrishnan, Yin, Molchanov, Yu, and Kautz]{huang2024lita}
De-An Huang, Shijia Liao, Subhashree Radhakrishnan, Hongxu Yin, Pavlo Molchanov, Zhiding Yu, and Jan Kautz.
\newblock {LITA}: Language instructed temporal-localization assistant.
\newblock In \emph{ECCV}, 2024.

\bibitem[Huang et~al.(2025)Huang, Radhakrishnan, Yu, and Kautz]{huang2025frag}
De-An Huang, Subhashree Radhakrishnan, Zhiding Yu, and Jan Kautz.
\newblock {FRAG}: Frame selection augmented generation for long video and long document understanding.
\newblock \emph{arXiv:2504.17447}, 2025.

\bibitem[Islam et~al.(2024)Islam, Ho, Yang, Nagarajan, Torresani, and Bertasius]{islam2024video}
Md~Mohaiminul Islam, Ngan Ho, Xitong Yang, Tushar Nagarajan, Lorenzo Torresani, and Gedas Bertasius.
\newblock {Video Recap}: Recursive captioning of hour-long videos.
\newblock In \emph{CVPR}, 2024.

\bibitem[Jin et~al.(2024)Jin, Sun, Xu, Chen, Jiang, Huang, Song, Liu, Zhang, Song, et~al.]{jin2024video}
Yang Jin, Zhicheng Sun, Kun Xu, Liwei Chen, Hao Jiang, Quzhe Huang, Chengru Song, Yuliang Liu, Di Zhang, Yang Song, et~al.
\newblock {Video-LaVIT}: Unified video-language pre-training with decoupled visual-motional tokenization.
\newblock \emph{arXiv:2402.03161}, 2024.

\bibitem[Kim et~al.(2024)Kim, Pertsch, Karamcheti, Xiao, Balakrishna, Nair, Rafailov, Foster, Lam, Sanketi, et~al.]{kim2024openvla}
Moo~Jin Kim, Karl Pertsch, Siddharth Karamcheti, Ted Xiao, Ashwin Balakrishna, Suraj Nair, Rafael Rafailov, Ethan Foster, Grace Lam, Pannag Sanketi, et~al.
\newblock {OpenVLA}: An open-source vision-language-action model.
\newblock \emph{arXiv:2406.09246}, 2024.

\bibitem[Lei et~al.(2021)Lei, Berg, and Bansal]{lei2021detecting}
Jie Lei, Tamara~L Berg, and Mohit Bansal.
\newblock Detecting moments and highlights in videos via natural language queries.
\newblock In \emph{NeurIPS}, 2021.

\bibitem[Li et~al.(2024{\natexlab{a}})Li, Zhang, Guo, Zhang, Li, Zhang, Zhang, Li, Liu, and Li]{li2024llavaonevision}
Bo Li, Yuanhan Zhang, Dong Guo, Renrui Zhang, Feng Li, Hao Zhang, Kaichen Zhang, Yanwei Li, Ziwei Liu, and Chunyuan Li.
\newblock {LLaVA-OneVision}: Easy visual task transfer.
\newblock \emph{arXiv:2408.03326}, 2024{\natexlab{a}}.

\bibitem[Li et~al.(2023)Li, Li, Savarese, and Hoi]{li2023blip}
Junnan Li, Dongxu Li, Silvio Savarese, and Steven Hoi.
\newblock {BLIP-2}: Bootstrapping language-image pre-training with frozen image encoders and large language models.
\newblock In \emph{ICML}, 2023.

\bibitem[Li et~al.(2024{\natexlab{b}})Li, Wang, He, Li, Wang, Liu, Wang, Xu, Chen, Luo, et~al.]{li2024mvbench}
Kunchang Li, Yali Wang, Yinan He, Yizhuo Li, Yi Wang, Yi Liu, Zun Wang, Jilan Xu, Guo Chen, Ping Luo, et~al.
\newblock {MVBench}: A comprehensive multi-modal video understanding benchmark.
\newblock In \emph{CVPR}, 2024{\natexlab{b}}.

\bibitem[Li et~al.(2024{\natexlab{c}})Li, Wang, and Jia]{li2024llama}
Yanwei Li, Chengyao Wang, and Jiaya Jia.
\newblock {LLaMA-VID}: An image is worth 2 tokens in large language models.
\newblock In \emph{ECCV}, 2024{\natexlab{c}}.

\bibitem[Li et~al.(2024{\natexlab{d}})Li, Chen, Han, Zhang, Wang, and Xie]{li2024multi}
Zeqian Li, Qirui Chen, Tengda Han, Ya Zhang, Yanfeng Wang, and Weidi Xie.
\newblock Multi-sentence grounding for long-term instructional video.
\newblock In \emph{ECCV}, 2024{\natexlab{d}}.

\bibitem[Lin et~al.(2024)Lin, Yin, Ping, Molchanov, Shoeybi, and Han]{lin2024vila}
Ji Lin, Hongxu Yin, Wei Ping, Pavlo Molchanov, Mohammad Shoeybi, and Song Han.
\newblock {VILA}: On pre-training for visual language models.
\newblock In \emph{CVPR}, 2024.

\bibitem[Liu et~al.(2024{\natexlab{a}})Liu, Yu, Lan, Wang, Fang, Kautz, Li, and Alvare]{liu2024streamchat}
Jihao Liu, Zhiding Yu, Shiyi Lan, Shihao Wang, Rongyao Fang, Jan Kautz, Hongsheng Li, and Jose~M Alvare.
\newblock {StreamChat}: Chatting with streaming video.
\newblock \emph{arXiv:2412.08646}, 2024{\natexlab{a}}.

\bibitem[Liu et~al.(2025{\natexlab{a}})Liu, Zhao, Xu, and Ghanem]{liu2025boltboostlargevisionlanguage}
Shuming Liu, Chen Zhao, Tianqi Xu, and Bernard Ghanem.
\newblock Bolt: Boost large vision-language model without training for long-form video understanding, 2025{\natexlab{a}}.

\bibitem[Liu et~al.(2025{\natexlab{b}})Liu, Wang, Ma, and Zhang]{liu2025video}
Xuyang Liu, Yiyu Wang, Junpeng Ma, and Linfeng Zhang.
\newblock Video compression commander: Plug-and-play inference acceleration for video large language models.
\newblock \emph{arXiv preprint arXiv:2505.14454}, 2025{\natexlab{b}}.

\bibitem[Liu et~al.(2024{\natexlab{b}})Liu, Ma, Qi, Wu, Shan, and Chen]{liu2024bench}
Ye Liu, Zongyang Ma, Zhongang Qi, Yang Wu, Ying Shan, and Chang~Wen Chen.
\newblock {E.T. Bench}: Towards open-ended event-level video-language understanding.
\newblock In \emph{NeurIPS}, 2024{\natexlab{b}}.

\bibitem[Liu et~al.(2025{\natexlab{c}})Liu, Dong, Liu, Hu, Lu, and Rao]{liu2025oryx}
Zuyan Liu, Yuhao Dong, Ziwei Liu, Winston Hu, Jiwen Lu, and Yongming Rao.
\newblock {Oryx MLLM}: On-demand spatial-temporal understanding at arbitrary resolution.
\newblock In \emph{ICLR}, 2025{\natexlab{c}}.

\bibitem[Luo et~al.(2025)Luo, Chen, Zheng, Huang, Yin, Lin, Fu, Huang, Ji, Luo, et~al.]{luo2025quota}
Yongdong Luo, Wang Chen, Xiawu Zheng, Weizhong Huang, Shukang Yin, Haojia Lin, Chaoyou Fu, Jinfa Huang, Jiayi Ji, Jiebo Luo, et~al.
\newblock Quota: Query-oriented token assignment via cot query decouple for long video comprehension.
\newblock \emph{arXiv preprint arXiv:2503.08689}, 2025.

\bibitem[Maaz et~al.(2024)Maaz, Rasheed, Khan, and Khan]{maaz2024video}
Muhammad Maaz, Hanoona Rasheed, Salman Khan, and Fahad~Shahbaz Khan.
\newblock {Video-ChatGPT}: Towards detailed video understanding via large vision and language models.
\newblock In \emph{ACL}, 2024.

\bibitem[Mangalam et~al.(2024)Mangalam, Akshulakov, and Malik]{mangalam2024egoschema}
Karttikeya Mangalam, Raiymbek Akshulakov, and Jitendra Malik.
\newblock {EgoSchema}: A diagnostic benchmark for very long-form video language understanding.
\newblock In \emph{NeurIPS}, 2024.

\bibitem[Meng et~al.(2022)Meng, Li, Chen, Lan, Wu, Jiang, and Lim]{meng2022adavit}
Lingchen Meng, Hengduo Li, Bor-Chun Chen, Shiyi Lan, Zuxuan Wu, Yu-Gang Jiang, and Ser-Nam Lim.
\newblock Adavit: Adaptive vision transformers for efficient image recognition.
\newblock In \emph{Proceedings of the IEEE/CVF conference on computer vision and pattern recognition}, pages 12309--12318, 2022.

\bibitem[Oncescu et~al.(2021)Oncescu, Henriques, Liu, Zisserman, and Albanie]{oncescu2021queryd}
Andreea-Maria Oncescu, Joao~F Henriques, Yang Liu, Andrew Zisserman, and Samuel Albanie.
\newblock {QuerYD}: A video dataset with high-quality text and audio narrations.
\newblock In \emph{ICASSP}, 2021.

\bibitem[OpenAI(2024)]{openai2024gpt4o}
OpenAI.
\newblock Hello gpt-4o.
\newblock \url{https://openai.com/index/hello-gpt-4o/}, 2024.

\bibitem[Pătrăucean et~al.(2023)Pătrăucean, Smaira, Gupta, Continente, Markeeva, Banarse, Koppula, Heyward, Malinowski, Yang, Doersch, Matejovicova, Sulsky, Miech, Frechette, Klimczak, Koster, Zhang, Winkler, Aytar, Osindero, Damen, Zisserman, and Carreira]{patraucean2023perception}
Viorica Pătrăucean, Lucas Smaira, Ankush Gupta, Adrià~Recasens Continente, Larisa Markeeva, Dylan Banarse, Skanda Koppula, Joseph Heyward, Mateusz Malinowski, Yi Yang, Carl Doersch, Tatiana Matejovicova, Yury Sulsky, Antoine Miech, Alex Frechette, Hanna Klimczak, Raphael Koster, Junlin Zhang, Stephanie Winkler, Yusuf Aytar, Simon Osindero, Dima Damen, Andrew Zisserman, and João Carreira.
\newblock {Perception Test}: A diagnostic benchmark for multimodal video models.
\newblock In \emph{NeurIPS}, 2023.

\bibitem[Qian et~al.(2024)Qian, Li, Wu, Ye, Fei, Chua, Zhuang, and Tang]{qian2024momentor}
Long Qian, Juncheng Li, Yu Wu, Yaobo Ye, Hao Fei, Tat-Seng Chua, Yueting Zhuang, and Siliang Tang.
\newblock Momentor: Advancing video large language model with fine-grained temporal reasoning.
\newblock In \emph{ICML}, 2024.

\bibitem[Ren et~al.(2024)Ren, Yao, Li, Sun, and Hou]{ren2024timechat}
Shuhuai Ren, Linli Yao, Shicheng Li, Xu Sun, and Lu Hou.
\newblock {TimeChat}: A time-sensitive multimodal large language model for long video understanding.
\newblock In \emph{CVPR}, 2024.

\bibitem[Shen et~al.(2024)Shen, Xiong, Zhao, Wu, Chen, Zhu, Liu, Xiao, Varadarajan, Bordes, Liu, Xu, J.~Kim, Soran, Krishnamoorthi, Elhoseiny, and Chandra]{shen2024longvu}
Xiaoqian Shen, Yunyang Xiong, Changsheng Zhao, Lemeng Wu, Jun Chen, Chenchen Zhu, Zechun Liu, Fanyi Xiao, Balakrishnan Varadarajan, Florian Bordes, Zhuang Liu, Hu Xu, Hyunwoo J.~Kim, Bilge Soran, Raghuraman Krishnamoorthi, Mohamed Elhoseiny, and Vikas Chandra.
\newblock {LongVU}: Spatiotemporal adaptive compression for long video-language understanding.
\newblock \emph{arXiv:2410.17434}, 2024.

\bibitem[Shu et~al.(2025)Shu, Liu, Zhang, Qin, Zhou, Liang, Huang, and Zhao]{shu2025video}
Yan Shu, Zheng Liu, Peitian Zhang, Minghao Qin, Junjie Zhou, Zhengyang Liang, Tiejun Huang, and Bo Zhao.
\newblock {Video-XL}: Extra-long vision language model for hour-scale video understanding.
\newblock In \emph{CVPR}, 2025.

\bibitem[Song et~al.(2024)Song, Chai, Wang, Zhang, Zhou, Wu, Chi, Guo, Ye, Zhang, et~al.]{song2024moviechat}
Enxin Song, Wenhao Chai, Guanhong Wang, Yucheng Zhang, Haoyang Zhou, Feiyang Wu, Haozhe Chi, Xun Guo, Tian Ye, Yanting Zhang, et~al.
\newblock {MovieChat}: From dense token to sparse memory for long video understanding.
\newblock In \emph{CVPR}, 2024.

\bibitem[Tang et~al.(2025)Tang, Qiu, Xie, Tian, Jiao, and Ye]{tang2025adaptive}
Xi Tang, Jihao Qiu, Lingxi Xie, Yunjie Tian, Jianbin Jiao, and Qixiang Ye.
\newblock Adaptive keyframe sampling for long video understanding.
\newblock In \emph{Proceedings of the Computer Vision and Pattern Recognition Conference}, pages 29118--29128, 2025.

\bibitem[Team et~al.(2023)Team, Anil, Borgeaud, Wu, Alayrac, Yu, Soricut, Schalkwyk, Dai, Hauth, et~al.]{team2023gemini}
Gemini Team, Rohan Anil, Sebastian Borgeaud, Yonghui Wu, Jean-Baptiste Alayrac, Jiahui Yu, Radu Soricut, Johan Schalkwyk, Andrew~M Dai, Anja Hauth, et~al.
\newblock Gemini: a family of highly capable multimodal models.
\newblock \emph{arXiv:2312.11805}, 2023.

\bibitem[Wang et~al.(2024{\natexlab{a}})Wang, Xu, Cheng, Diao, Zhou, Cao, Wang, Ge, and Huang]{wang2024grounded}
Haibo Wang, Zhiyang Xu, Yu Cheng, Shizhe Diao, Yufan Zhou, Yixin Cao, Qifan Wang, Weifeng Ge, and Lifu Huang.
\newblock {Grounded-VideoLLM}: Sharpening fine-grained temporal grounding in video large language models.
\newblock \emph{arXiv:2410.03290}, 2024{\natexlab{a}}.

\bibitem[Wang et~al.(2022)Wang, Yang, Li, Liu, Wu, and Jiang]{wang2022efficient}
Junke Wang, Xitong Yang, Hengduo Li, Li Liu, Zuxuan Wu, and Yu-Gang Jiang.
\newblock Efficient video transformers with spatial-temporal token selection.
\newblock In \emph{European Conference on Computer Vision}, pages 69--86. Springer, 2022.

\bibitem[Wang et~al.(2024{\natexlab{b}})Wang, Yuan, Zhang, and Sun]{wang2024tarsier}
Jiawei Wang, Liping Yuan, Yuchen Zhang, and Haomiao Sun.
\newblock Tarsier: Recipes for training and evaluating large video description models.
\newblock \emph{arXiv:2407.00634}, 2024{\natexlab{b}}.

\bibitem[Wang et~al.(2024{\natexlab{c}})Wang, Bai, Tan, Wang, Fan, Bai, Chen, Liu, Wang, Ge, et~al.]{wang2024qwen2}
Peng Wang, Shuai Bai, Sinan Tan, Shijie Wang, Zhihao Fan, Jinze Bai, Keqin Chen, Xuejing Liu, Jialin Wang, Wenbin Ge, et~al.
\newblock {Qwen2-VL}: Enhancing vision-language model's perception of the world at any resolution.
\newblock \emph{arXiv:2409.12191}, 2024{\natexlab{c}}.

\bibitem[Wang et~al.(2024{\natexlab{d}})Wang, Xie, Liu, and Zheng]{wang2024videollamb}
Yuxuan Wang, Cihang Xie, Yang Liu, and Zilong Zheng.
\newblock {VideoLLaMB}: Long-context video understanding with recurrent memory bridges.
\newblock \emph{arXiv:2409.01071}, 2024{\natexlab{d}}.

\bibitem[Wei and Chen(2024)]{wei2024visual}
Hongchen Wei and Zhenzhong Chen.
\newblock Visual context window extension: A new perspective for long video understanding.
\newblock \emph{arXiv:2409.20018}, 2024.

\bibitem[Xiao et~al.(2021)Xiao, Shang, Yao, and Chua]{xiao2021next}
Junbin Xiao, Xindi Shang, Angela Yao, and Tat-Seng Chua.
\newblock {NExT-QA}: Next phase of question-answering to explaining temporal actions.
\newblock In \emph{CVPR}, 2021.

\bibitem[Xu et~al.(2024{\natexlab{a}})Xu, Zhao, Zhou, Lin, Ng, and Feng]{xu2024pllava}
Lin Xu, Yilin Zhao, Daquan Zhou, Zhijie Lin, See~Kiong Ng, and Jiashi Feng.
\newblock {PLLaVA}: Parameter-free llava extension from images to videos for video dense captioning.
\newblock \emph{arXiv:2404.16994}, 2024{\natexlab{a}}.

\bibitem[Xu et~al.(2024{\natexlab{b}})Xu, Gao, Gan, Chen, Lai, Gang, Kang, and Dehghan]{xu2024slowfast}
Mingze Xu, Mingfei Gao, Zhe Gan, Hong-You Chen, Zhengfeng Lai, Haiming Gang, Kai Kang, and Afshin Dehghan.
\newblock {SlowFast-LLaVA}: A strong training-free baseline for video large language models.
\newblock \emph{arXiv:2407.15841}, 2024{\natexlab{b}}.

\bibitem[Yao et~al.(2025)Yao, Wu, Ouyang, Zhang, Xiong, Chen, Sun, and Li]{yao2025generative}
Linli Yao, Haoning Wu, Kun Ouyang, Yuanxing Zhang, Caiming Xiong, Bei Chen, Xu Sun, and Junnan Li.
\newblock Generative frame sampler for long video understanding.
\newblock \emph{arXiv preprint arXiv:2503.09146}, 2025.

\bibitem[Ye et~al.(2024)Ye, Xu, Liu, Hu, Yan, Qian, Zhang, Huang, and Zhou]{ye2024mplug}
Jiabo Ye, Haiyang Xu, Haowei Liu, Anwen Hu, Ming Yan, Qi Qian, Ji Zhang, Fei Huang, and Jingren Zhou.
\newblock {mPLUG-Owl3}: Towards long image-sequence understanding in multi-modal large language models.
\newblock In \emph{ICLR}, 2024.

\bibitem[Yu et~al.(2023)Yu, Cho, Yadav, and Bansal]{yu2023self}
Shoubin Yu, Jaemin Cho, Prateek Yadav, and Mohit Bansal.
\newblock Self-chained image-language model for video localization and question answering.
\newblock In \emph{NeurIPS}, 2023.

\bibitem[Yu et~al.(2025{\natexlab{a}})Yu, Jin, Wang, Chen, Jin, Zuo, Xu, Sun, Zhang, Wu, Zhang, and Sun]{yu2025framevoyagerlearningqueryframes}
Sicheng Yu, Chengkai Jin, Huanyu Wang, Zhenghao Chen, Sheng Jin, Zhongrong Zuo, Xiaolei Xu, Zhenbang Sun, Bingni Zhang, Jiawei Wu, Hao Zhang, and Qianru Sun.
\newblock Frame-voyager: Learning to query frames for video large language models, 2025{\natexlab{a}}.

\bibitem[Yu et~al.(2025{\natexlab{b}})Yu, Jin, Wang, Chen, Jin, Zuo, Xu, Sun, Zhang, Wu, et~al.]{yu2025frame}
Sicheng Yu, Chengkai Jin, Huanyu Wang, Zhenghao Chen, Sheng Jin, Zhongrong Zuo, Xiaolei Xu, Zhenbang Sun, Bingni Zhang, Jiawei Wu, et~al.
\newblock {Frame-Voyager}: Learning to query frames for video large language models.
\newblock In \emph{ICLR}, 2025{\natexlab{b}}.

\bibitem[Zala et~al.(2023)Zala, Cho, Kottur, Chen, Oguz, Mehdad, and Bansal]{zala2023hierarchical}
Abhay Zala, Jaemin Cho, Satwik Kottur, Xilun Chen, Barlas Oguz, Yashar Mehdad, and Mohit Bansal.
\newblock Hierarchical video-moment retrieval and step-captioning.
\newblock In \emph{CVPR}, 2023.

\bibitem[Zhai et~al.(2023)Zhai, Mustafa, Kolesnikov, and Beyer]{zhai2023sigmoid}
Xiaohua Zhai, Basil Mustafa, Alexander Kolesnikov, and Lucas Beyer.
\newblock Sigmoid loss for language image pre-training.
\newblock In \emph{ICCV}, 2023.

\bibitem[Zhang et~al.(2024{\natexlab{a}})Zhang, Wang, Tang, Liu, Feng, Dai, and Jin]{zhang2024flash}
Haoji Zhang, Yiqin Wang, Yansong Tang, Yong Liu, Jiashi Feng, Jifeng Dai, and Xiaojie Jin.
\newblock {Flash-VStream}: Memory-based real-time understanding for long video streams.
\newblock \emph{arXiv:2406.08085}, 2024{\natexlab{a}}.

\bibitem[Zhang et~al.(2024{\natexlab{b}})Zhang, Li, Zhang, Pu, Cahyono, Hu, Liu, Zhang, Yang, Li, et~al.]{zhang2024lmms}
Kaichen Zhang, Bo Li, Peiyuan Zhang, Fanyi Pu, Joshua~Adrian Cahyono, Kairui Hu, Shuai Liu, Yuanhan Zhang, Jingkang Yang, Chunyuan Li, et~al.
\newblock {LMMs-Eval}: Reality check on the evaluation of large multimodal models.
\newblock \emph{arXiv:2407.12772}, 2024{\natexlab{b}}.

\bibitem[Zhang et~al.(2024{\natexlab{c}})Zhang, Dong, Zang, Cao, Qian, Chen, Guo, Duan, Wang, Ouyang, et~al.]{zhang2024internlm}
Pan Zhang, Xiaoyi Dong, Yuhang Zang, Yuhang Cao, Rui Qian, Lin Chen, Qipeng Guo, Haodong Duan, Bin Wang, Linke Ouyang, et~al.
\newblock {InternLM-XComposer-2.5}: A versatile large vision language model supporting long-contextual input and output.
\newblock \emph{arXiv:2407.03320}, 2024{\natexlab{c}}.

\bibitem[Zhang et~al.(2024{\natexlab{d}})Zhang, Zhang, Li, Zeng, Yang, Zhang, Wang, Tan, Li, and Liu]{zhang2024long}
Peiyuan Zhang, Kaichen Zhang, Bo Li, Guangtao Zeng, Jingkang Yang, Yuanhan Zhang, Ziyue Wang, Haoran Tan, Chunyuan Li, and Ziwei Liu.
\newblock Long context transfer from language to vision.
\newblock \emph{arXiv:2406.16852}, 2024{\natexlab{d}}.

\bibitem[Zhang et~al.(2025)Zhang, Yang, Yin, Luo, and Luan]{zhang2025q}
Shaojie Zhang, Jiahui Yang, Jianqin Yin, Zhenbo Luo, and Jian Luan.
\newblock Q-frame: Query-aware frame selection and multi-resolution adaptation for video-llms.
\newblock \emph{arXiv preprint arXiv:2506.22139}, 2025.

\bibitem[Zhang et~al.(2024{\natexlab{e}})Zhang, Wu, Li, Li, Ma, Liu, and Li]{zhang2024video}
Yuanhan Zhang, Jinming Wu, Wei Li, Bo Li, Zejun Ma, Ziwei Liu, and Chunyuan Li.
\newblock Video instruction tuning with synthetic data.
\newblock \emph{arXiv:2410.02713}, 2024{\natexlab{e}}.

\bibitem[Zhou et~al.(2024{\natexlab{a}})Zhou, Shu, Zhao, Wu, Xiao, Yang, Xiong, Zhang, Huang, and Liu]{zhou2024mlvu}
Junjie Zhou, Yan Shu, Bo Zhao, Boya Wu, Shitao Xiao, Xi Yang, Yongping Xiong, Bo Zhang, Tiejun Huang, and Zheng Liu.
\newblock {MLVU}: A comprehensive benchmark for multi-task long video understanding.
\newblock \emph{arXiv:2406.04264}, 2024{\natexlab{a}}.

\bibitem[Zhou et~al.(2024{\natexlab{b}})Zhou, Arnab, Buch, Yan, Myers, Xiong, Nagrani, and Schmid]{zhou2024streaming}
Xingyi Zhou, Anurag Arnab, Shyamal Buch, Shen Yan, Austin Myers, Xuehan Xiong, Arsha Nagrani, and Cordelia Schmid.
\newblock Streaming dense video captioning.
\newblock In \emph{CVPR}, 2024{\natexlab{b}}.

\bibitem[Zohar et~al.(2024)Zohar, Wang, Dubois, Mehta, Xiao, Hansen-Estruch, Yu, Wang, Juefei-Xu, Zhang, et~al.]{zohar2024apollo}
Orr Zohar, Xiaohan Wang, Yann Dubois, Nikhil Mehta, Tong Xiao, Philippe Hansen-Estruch, Licheng Yu, Xiaofang Wang, Felix Juefei-Xu, Ning Zhang, et~al.
\newblock Apollo: An exploration of video understanding in large multimodal models.
\newblock \emph{arXiv:2412.10360}, 2024.

\end{thebibliography}
}

\appendix
\clearpage
\noindent
\begin{center}
    \Large \textbf{Appendix}
\end{center}
\vspace{-1em} 

\section{The Use of Large Language Models (LLMs)}

\begin{algorithm}[h]
\caption{Keyframe Extraction via Bidirectional CLIP Similarity}
\begin{algorithmic}[1]
\REQUIRE Video frame sequence \texttt{frames}, similarity thresholds $t_1$ (scene change) and $t_2$ (diversity)
\ENSURE Selected keyframe indices \texttt{sel}
\STATE Initialize \texttt{sel} with the first frame index: \texttt{sel} $\leftarrow \{0\}$
\STATE Extract CLIP feature for the first frame: \texttt{prev} $\leftarrow$ clip(\texttt{frames}[0])
\FOR{each frame $c$ in \texttt{frames[1:]} with index $i$}
    \STATE \texttt{curr} $\leftarrow$ clip($c$)
    \STATE $s \leftarrow$ sim(\texttt{curr}, \texttt{prev})
    \IF{$s < t_1$}
        \FOR{each future frame $f$ in \texttt{frames}[$i+1$ : ]}
            \STATE \texttt{fut} $\leftarrow$ clip($f$)
            \IF{sim(\texttt{curr}, \texttt{fut}) $< t_2$}
                \STATE Add index $i$ to \texttt{sel}
                \STATE \texttt{prev} $\leftarrow$ \texttt{curr}
                \STATE \textbf{break}
            \ENDIF
        \ENDFOR
    \ENDIF
\ENDFOR
\IF{sim(clip(\texttt{frames}[-1]), \texttt{prev}) $< t_1$}
    \STATE Add last frame index to \texttt{sel}
\ENDIF
\RETURN \texttt{sel}
\end{algorithmic}
\label{alg:keyframe_clip}
\end{algorithm}

In this work, Large Language Models (LLMs) were employed in four main ways: (i) to aid and polish the writing for clarity and style; and (ii) to provide coding assistance, including code generation, debugging, and optimization suggestions.

Specifically, LLMs were utilized to improve the clarity, coherence, and readability of the manuscript, with particular attention given to the \textbf{Related Work}, \textbf{Method}, and \textbf{Experiments} sections. In these parts, the initial drafts were carefully reviewed and refined using LLM-powered suggestions for sentence structure, terminology, and logical flow. This process ensured that the technical content was presented in a precise and accessible manner, while maintaining consistency in academic tone and style throughout the paper.

All outputs generated by LLMs were critically reviewed, verified, and further refined by the authors. The core scientific ideas, methodology, and contributions remain entirely the authors’ own. The use of LLMs was strictly limited to language enhancement and coding support, without influencing the originality or integrity of the research.

\section{Inference time}
In Table~\ref{tab:table1}, we evaluated the speed of our model on a single NVIDIA A100 GPU. We employed LLaVA-Video-7B~\cite{zhang2024video} as our answering LLM, implemented a 32-frame sampling strategy from 512 input frames in total, and generated 27 text tokens. Additionally, we leveraged KV Cache and Flash Attention~\cite{dao2022flashattention, dao2023flashattention2} to enhance inference efficiency.

Our detailed analysis of computational costs reveals that processing each video sample requires a total of $6.42$ seconds, with the Vision Encoder ($2.92$ seconds) and LLM ($2.89$ seconds) dominating the time consumption. These two components collectively consume $90\%$ of the total processing time, indicating the direction for future system optimization. In contrast, our VideoITG module demonstrates remarkable efficiency, requiring only $0.61$ seconds to scan $512$ frames—a speed that surpasses human visual recognition and thinking capabilities.

\begin{table}[ht]
  \centering
  \setlength{\tabcolsep}{1.5mm}{
  \caption{Computation cost of the model.}\vspace{2mm}
  \label{tab:table1}
  \begin{tabular}{c c c c c c}
  \toprule
   Vision Encoder & VideoITG & LLM & Overall \\
  \midrule
   2.92s & 0.61s & 2.89s & 6.42s  \\
  \bottomrule
  \end{tabular}
  } 
\end{table}

\begin{table*}[!h]
\centering
\caption{The performance (accuracy) of SOTA methods on video benchmarks. For InternVL2.5-8B results, we report the higher results in the technical report and lmms-eval. We sample 32 frames using VideoITG for our results.}\vspace{1mm}
\label{tab:video-bench}
\renewcommand{\arraystretch}{0.9}
\small
\resizebox{\textwidth}{!}{
\setlength{\tabcolsep}{1pt}
\begin{tabular}{@{}lc|ccccccc@{}}
    \toprule
     & \multicolumn{1}{c|}{\scriptsize{\textbf{Open-Ended Q\&A}}} & \multicolumn{7}{c}{\scriptsize{\textbf{Multi-Choice Q\&A}}}  \\   
    \multirow{2}{*}{\textbf{Model}} & \rotatebox{45}{\textbf{\scriptsize{ActNet-QA}}} & \rotatebox{45}{\textbf{\scriptsize{EgoSchema}}} & \rotatebox{45}{\textbf{\scriptsize{MLVU}}} & \rotatebox{45}{\textbf{\scriptsize{NExT-QA}}} & \rotatebox{45}{\textbf{\scriptsize{PerceptionTest}}}  & \rotatebox{45}{\textbf{\scriptsize{LongVideoBench}}} &\rotatebox{45}{\textbf{\scriptsize{VideoMME}}} &\rotatebox{45}{\textbf{\scriptsize{MVBench}}} \\ \cmidrule(l){2-9} 
    & test & test & m-avg & mc & val   & val & wo/w-subs & val \\ \midrule
    \multicolumn{9}{l}{\textit{Open-source models}} \\
    VILA-40B~\cite{lin2024vila} & 58.0 & 58.0 & - & 67.9 & 54.0  & -  & 60.1/61.1 & - \\
    PLLaVA-34B~\cite{xu2024pllava} & 60.9 & - & - & - & - & 53.2 & - & 58.1 \\    
    VideoLLaMA2-7B~\cite{cheng2024videollama} & 50.2 & 50.5 & - & - & 49.6 & - & 45.1/46.6 & 53.4 \\ 
    LongVA-7B~\cite{zhang2024long} & 50.0 & - & 56.3 & 68.3 & -  & -  & 52.6/54.3 & - \\ 
    LongVU-7B~\cite{zhang2024long} & - & 67.6 & 65.4 & - & -  & -  & 60.6/- & 66.9 \\
    LLaVA-OV-7B~\cite{li2024llavaonevision} & 56.6 & 60.1 & 64.7 & 79.4 & 57.1    & 56.5 & 58.2/61.5 & 56.7 \\
    mPLUG-Owl3-8B~\cite{ye2024mplug} & - & - & - & 78.6 & - & 52.1 & 53.5/- & 54.5 \\ 
    LLaVA-Video-7B~\cite{zhang2024video} & 56.5 & 57.3 & 70.8 & 83.2 & 67.9 & 58.2 & 63.3/69.7 & 58.6 \\
    Qwen2.5-VL-7B~\cite{wang2024qwen2} & - & - & - & 70.2 & 70.5 & 54.7 & 65.1/71.6 & 69.6 \\
    InternVL2.5-8B~\cite{zhang2024internlm} & - & 51.5 & 68.9 & - & - & 60.0 & 64.2/66.9 & 72.0 \\
    \midrule 
    \rowcolor[HTML]{DAEFF9}
    InternVL2.5-8B-ITG-32 & 57.4 & 51.6  & 75.0 & 79.5 & 64.9 & 61.9 & 67.3/69.6 & 72.2 \\ 
    \bottomrule
    \end{tabular}
}
\end{table*}

\begin{table*}[t]
\centering
\caption{Dataset quality (IoU). We evaluate the performance in both multiple-choice (MC) and open-ended (OE) questions.}
\label{tab:ablate_dataset}\vspace{2mm}
\resizebox{0.99\textwidth}{!}{
\setlength{\tabcolsep}{3.5pt}
\begin{tabular}{c|cccc}  
\toprule
\textbf{Method} & \textbf{Semantic-MC} & \textbf{Semantic \& Motion-OE} & \textbf{Semantic-MC} & \textbf{Semantic \& Motion-OE}  \\
\midrule
Qwen2.5-VL-32B & 0.31 & 0.36 & 0.27 & 0.37 \\
GPT4o & 0.24 & 0.30 & 0.26 & 0.27 \\
Ours & \textbf{0.79} & \textbf{0.74} & \textbf{0.72} & \textbf{0.69} \\
\bottomrule
\end{tabular}
}
\end{table*}

\section{Dataset details}
\subsection{Prompt template}
Our Question-guided Clip Retrieval process utilizes a carefully designed prompt template (shown in Table~\ref{tab:temporal_grounding_prompt}) that instructs the LLM to analyze chronologically ordered clip-level descriptions and identify the minimal set of clips necessary to answer a given question.
The prompt template consists of three main components:

\begin{itemize}[leftmargin=6mm]
    \item \textbf{Task Description}: Defines the LLM's role as an expert in analyzing video clip descriptions and establishes the goal of selecting clips that cover both question and answer content.
    \item \textbf{Guidelines}: Provides detailed instructions for clip selection, including handling time-related expressions, determining if a single or multiple clips are needed, addressing questions about object existence or movement, and avoiding unnecessary clips.
    \item \textbf{Output Format}: Specifies the required JSON structure for responses, ensuring consistent formatting with explanation and clip number fields.
\end{itemize}

This template enables the LLM to perform chain-of-thought reasoning when selecting relevant clips. The model analyzes keywords from questions, identifies temporal relationships (\eg, ``before," ``after"), and provides explicit rationales for its selections. For cases where no relevant clips exist, the model returns ``None" to reduce annotation noise.

\begin{table*}[t!]\centering
\caption{Prompt Template: An expert system for temporal localization in video segments. The system analyzes video segment descriptions to determine the minimal and necessary combination of segments required to answer questions.}
\label{tab:temporal_grounding_prompt}\vspace{2mm}
\begin{minipage}{\linewidth}
\centering
\begin{tcolorbox} 
\centering
\footnotesize
\begin{tabular}{p{0.9\columnwidth} c}
\normalsize\color{blue}{\textbf{Task:}} &\\
You are an expert in analyzing video clip descriptions. Your task is to select which clip or combination of clips is necessary to answer the given question, ensuring the selected clips effectively cover the content of both the question and the answer. & \\
\midrule
\normalsize\color{blue}{\textbf{Guidelines:}} & \\
\begin{itemize}[leftmargin=6mm]
\item Carefully read the descriptions to determine which clip(s) provide relevant content for the question and the answer.
\item Clip descriptions are in chronological order. Use clip number to locate clips based on time-related expressions (e.g., "at the beginning of the video" suggests a smaller clip number, while "at the end of the video" suggests a larger one).
\item First, determine if one clip can answer the question or if multiple clips are needed. Then, return a list containing the selected clip(s) and an explanation.
\item If the question asks about the existence/movement of an object or event. The object/action/movement may not exist, meaning you can't find the answer in the description, but the question might still provide some clues. You need to find the sentence closest to those clues.
\item If asked about the whole video description or overall atmosphere, you should return all clip numbers.
\item If multiple clips provide similar descriptions of the content and any of them can be used to answer the question, return all corresponding clips.
\item If there are no clues in all descriptions and cannot answer the question, return "None.".
\item \textbf{Important}: Avoid including unnecessary clips.
\end{itemize} \\
\midrule

\normalsize\color{blue}{\textbf{Output Format:}} & \\
1. Your output should be formed in a JSON file. \\
2. Only return the Python dictionary string. \\
For example: \\
{\small\bfseries\texttt{\{"explanation": "...", "clip\_num": "One clip: [Clip-2]"\}}} \\
{\small\bfseries\texttt{\{"explanation": "...", "clip\_num": "Multiple clips: [Clip-1, Clip-7, Clip-8]"\}}} \\
{\small\bfseries\texttt{\{"explanation": "...", "clip\_num": "None."\}}} \\
    \end{tabular}
\end{tcolorbox}
\vspace{-2mm}
\end{minipage}
\end{table*}

\begin{table*}[t!]\centering
\caption{Prompt template for identifying motion-related questions in video QA tasks. The template instructs the system to analyze each question-answer pair and determine whether the question pertains to absolute or relative speed, responding with ``Yes'' or ``No'' accordingly. Example cases are provided for clarification.}
\label{tab:temporal_grounding_prompt_v1}\vspace{2mm}
\begin{minipage}{\linewidth}
\centering
\begin{tcolorbox} 
\centering
\footnotesize
\begin{tabular}{p{0.9\columnwidth} c}
\normalsize\color{blue}{\textbf{Task:}} &\\
Analyze the given QA pair to determine if the question is related to speed. Specifically, check if it involves either absolute speed (the speed of a specific object) or relative speed (comparing the speed of different objects). Provide an output of "Yes" if the question pertains to speed, and "No" otherwise. & \\
\textbf{Important}: Respond with "Yes" or "No" only.  & \\
\midrule
\normalsize\color{blue}{\textbf{Example:}} & \\
\textbf{Question 1:} Which is faster, the white car or the bicycle? Options: 
A. The bicycle. 
B. The white car. 
C. Both are at the same speed. 
D. None of the above. \\
\textbf{Answer 1:} B. The white car. \\
\textbf{Output:} Yes. \\
\textbf{Question 2:} What color is the cat ?Options: 
A. black 
B. white 
C. orange 
D. gray  \\
\textbf{Answer 2:} C. orange  \\
\textbf{Output:} No. \\
    \end{tabular}
\end{tcolorbox}
\vspace{-2mm}
\end{minipage}
\end{table*}

\begin{table*}[t!]\centering
\caption{Prompt template for identifying semantic-related questions in video QA tasks. The template instructs the system to analyze each question-answer pair and determine whether the question pertains to absolute or relative speed, responding with ``Yes'' or ``No'' accordingly. Example cases are provided for clarification.}
\label{tab:temporal_grounding_prompt_v2}\vspace{2mm}
\begin{minipage}{\linewidth}\vspace{0mm}
\centering
\begin{tcolorbox} 
\centering
\footnotesize
\begin{tabular}{p{0.9\columnwidth} c}
\normalsize\color{blue}{\textbf{Task:}} &\\
Analyze the given QA pair to determine if the question inquires about the existence of an object or action. If it does, and the answer is "No" (indicating non-existence), output "Yes." If the question is not about existence, or the answer is "Yes" (indicating existence), output "No."  & \\
\textbf{Important}: Respond with "Yes" or "No" only.  & \\
\midrule
\normalsize\color{blue}{\textbf{Example:}} & \\
\textbf{Question 1:} After going through the bag, does the person meticulously clean the area around the sink?  \\
\textbf{Answer 1:} No, the person does not clean the area around the sink after going through the bag. The video primarily focuses on the action of the person with the bag and items, not on cleaning activities. \\
\textbf{Output:} Yes. \\
\textbf{Question 2:} Is there a cat sitting on the windowsill in the video?  \\
\textbf{Answer 2:} Yes, there is a cat sitting on the windowsill throughout the video. \\
\textbf{Output:} No. \\
\midrule

\end{tabular}
\end{tcolorbox}
\vspace{-2mm}
\end{minipage}
\end{table*}

We implement this process using GPT-4o-mini~\cite{openai2024gpt4o}, which is sufficient for accurate clip selection while reducing annotation costs by over 10 times compared to larger models. The selected clips are then converted to event boundaries defined by timestamps based on frame indices for the final temporal grounding annotations.

\subsection{Human-in-the-loop verification}

Ensuring the quality of automatically annotated datasets is critical for the reliability and effectiveness of downstream video understanding models. In this work, we implement a comprehensive quality control protocol for the VideoITG-40K dataset.

Our pipeline begins with diverse sampling: we select a representative subset of the dataset, covering a wide range of instructions and video scenarios. For this subset, we conduct human verification, where expert annotators review the automatically generated annotations to assess their accuracy and relevance. This process allows us to identify and correct potential errors, and to further calibrate our annotation pipeline for improved consistency and quality.

As shown in Table~\ref{tab:ablate_dataset}, we compare our pipeline with baselines where advanced models such as Qwen2.5VL and GPT-4o are directly prompted to answer the temporal boundaries of relevant events. These direct approaches result in significantly lower performance, highlighting the advantage of our multi-step, instruction-guided annotation strategy.

\subsection{Frame Sampling Algorithm}
The algorithm in ~\ref{alg:keyframe_clip} is designed for semantic-only keyframe selection, aiming to extract a diverse set of frames that comprehensively capture the semantic content of a video—such as people, scenes, or objects. By leveraging CLIP features, the algorithm compares each frame to previously selected keyframes using a bidirectional similarity measure. Frames are selected when their semantic features differ significantly from both the last keyframe (scene change threshold) and from future frames (diversity threshold), ensuring that each chosen frame represents distinct semantic information. This process produces a set of keyframes with maximal semantic coverage and minimal redundancy, aligning with the goal of representing all major semantic aspects of the video.

\section{Visualization}
\begin{figure*}[ht]
\centering
\includegraphics[width=\linewidth]{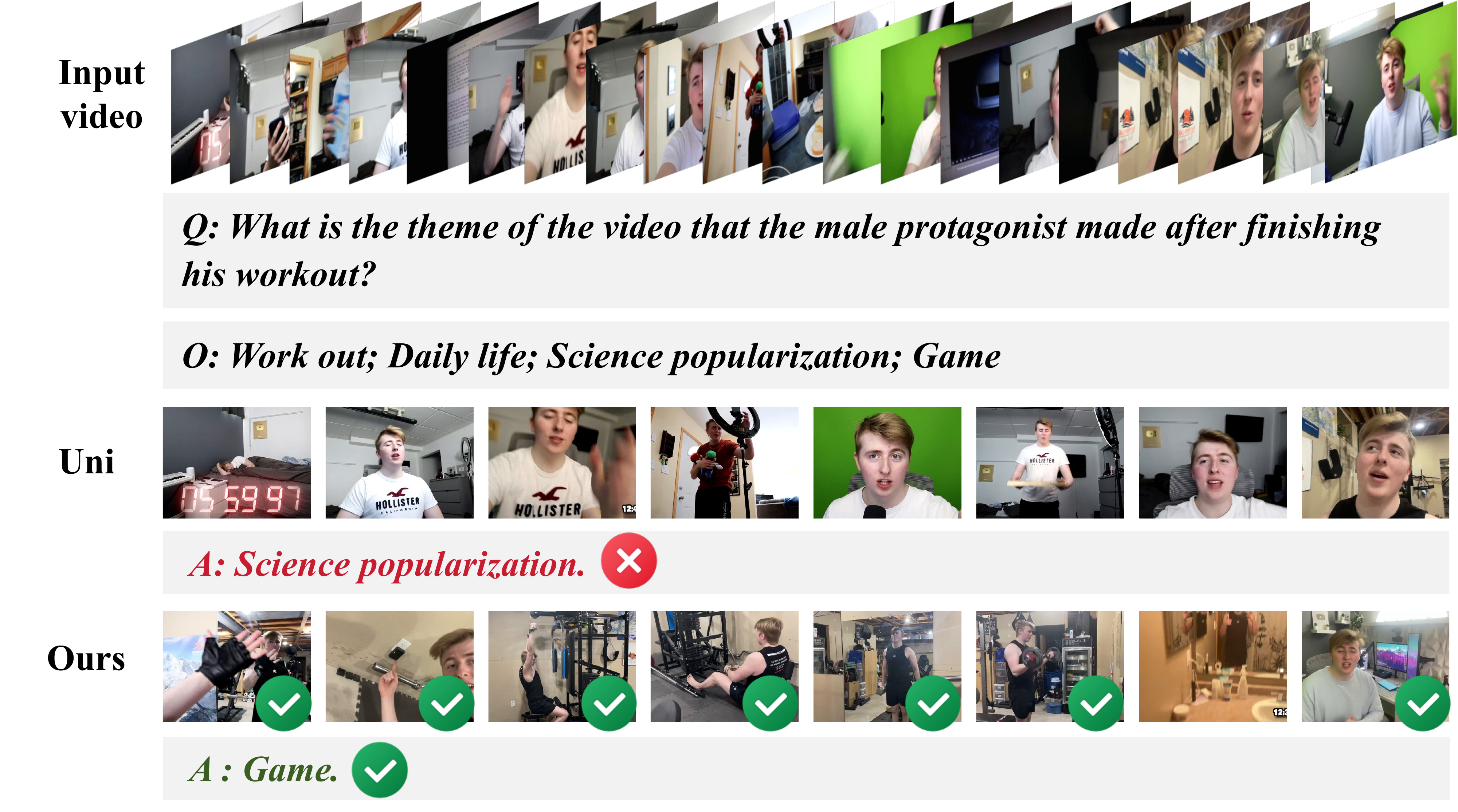}
\caption{Example-1 shows how different sampling strategies impact video understanding. We mark the identified key frames that directly answer the question with green check-marks.}
\label{vis_a1}
\end{figure*}

\begin{figure*}[!htbp]
\centering
\includegraphics[width=\linewidth]{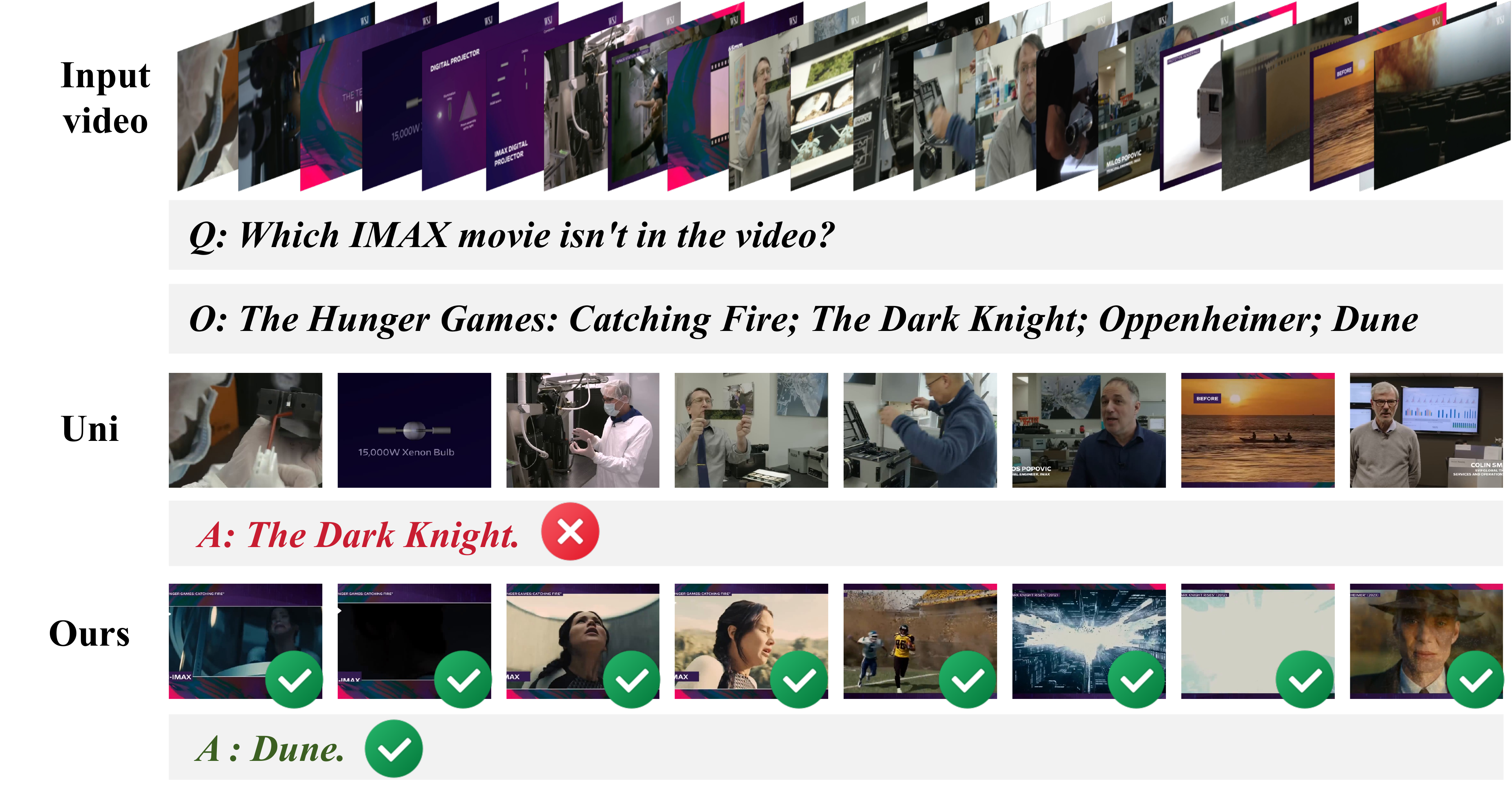}
\caption{Example-2 shows how different sampling strategies impact video understanding. We mark the identified key frames that directly answer the question with green check-marks.}
\label{vis_a2}
\end{figure*}

In Fig.~\ref{vis_a1} and Fig.~\ref{vis_a2}, 
we present two sets of results comparing sampling results of VideoITG with uniform sampling. Fig.~\ref{vis_a1} demonstrates a temporal reasoning problem, where our model accurately identifies the ``workout'' mentioned in the question and successfully locates the subsequent actions in the video, leading to the correct answer selection. In contrast, the uniform sampling strategy failed to capture these crucial frames. 
Fig.~\ref{vis_a2} illustrates a non-existence question scenario where our model effectively identifies all IMAX movies present in the given options, enabling it to successfully filter out and determine the correct answer.

\clearpage

\end{document}